\documentclass[]{opendatalab}
\usepackage{xcolor}
\usepackage{longtable}
\usepackage{listings}
\usepackage[numbers]{natbib}
\definecolor{codegreen}{rgb}{0,0.6,0}
\definecolor{codegray}{rgb}{0.5,0.5,0.5}
\definecolor{codepurple}{rgb}{0.58,0,0.82}
\definecolor{backcolour}{rgb}{0.95,0.95,0.92}
\definecolor{promptcolor}{HTML}{D1D0F2}
\definecolor{promptcolorheader}{HTML}{bdbcec}
\newcommand{\promptbox}[2]{
\begin{tcolorbox}[
top=0.3em,bottom=0.3em,left=0.5em,right=0.5em,
toptitle=0.3em,bottomtitle=0.2em,boxsep=0pt,
colframe=promptcolorheader,colback=promptcolor!50,boxrule=0.5pt,
]
\footnotesize
\end{tcolorbox}
}
\lstdefinestyle{mystyle}{
    backgroundcolor=\color{backcolour},   
    commentstyle=\color{codegreen},
    keywordstyle=\color{magenta},
    numberstyle=\tiny\color{codegray},
    stringstyle=\color{codepurple},
    basicstyle=\ttfamily\footnotesize,
    breakatwhitespace=false,         
    breaklines=true,                 
    captionpos=b,                    
    keepspaces=true,                 
    numbers=left,                    
    numbersep=5pt,                  
    showspaces=false,                
    showstringspaces=false,
    showtabs=false,                  
    tabsize=2
}

\usepackage{algorithm}
\usepackage{algorithmic}
\usepackage{multirow}
\usepackage{booktabs}
\usepackage{enumitem}
\usepackage{tabularx}
\usepackage{amsmath}
\usepackage{amssymb}
\usepackage{graphicx}
\usepackage{listings}
\usepackage{xcolor}
\usepackage{multirow}
\usepackage{tcolorbox}
\tcbuselibrary{skins}
\usepackage[table,xcdraw]{xcolor}
\usepackage{makecell}
\usepackage{enumitem}
\usepackage{mhchem}

\usepackage[T1]{fontenc}
\usepackage[utf8]{inputenc}

\usepackage{pifont}        

\usepackage{listings}
\newcommand{\code}[1]{\lstinline!#1!}
\lstdefinestyle{prompt}{
    basicstyle=\ttfamily\fontsize{7pt}{8pt}\selectfont,
    frame=none,
    breaklines=true,
    backgroundcolor=\color{lightgray},
    breakatwhitespace=true,
    breakindent=0pt,
    escapeinside={(*@}{@*)},
    numbers=none,
    numbersep=5pt,
    xleftmargin=5pt,
    literate={`}{\textasciigrave}1
}
\tcbset{
  aibox/.style={
    top=10pt,
    colback=white,
    colframe=black,
    colbacktitle=black,
    enhanced,
    center,
    attach boxed title to top left={yshift=-0.1in,xshift=0.15in},
    boxed title style={boxrule=0pt,colframe=white,},
  }
}
\newtcolorbox{AIbox}[2][]{aibox, title=#2,#1}
\lstdefinestyle{json}{
    basicstyle=\ttfamily\small,       
    stringstyle=\color{brown},        
    keywordstyle=\color{blue},        
    numberstyle=\color{magenta},      
    commentstyle=\color{green!50!black}, 
    morestring=[b]",                  
    morekeywords={true,false,null},   
    breaklines=true,                  
    xleftmargin=10pt,                 
    xrightmargin=10pt,                
}

\usepackage{newfloat}
\usepackage{listings}

\title{MolRecBench-Wild: A Real-World Benchmark for Optical Chemical Structure Recognition}

\author[1\dag]{Haote Yang}
\author[2\dag]{Hui Wang}
\author[3\dag]{Chen Zhu}
\author[4\dag]{Jingchao Wang}
\author[5\dag]{Linye Li}
\author[6\dag]{Hongbin Lai}
\author[7\dag]{Huijie Ao}
\author[8\dag]{Yongxuan Lyu}
\author[1\dag]{Jiang Wu}
\author[1]{Jiaxing Sun}
\author[1]{Lua Chen}
\author[1]{Yuanyuan Cao}
\author[1]{Ruijie Zhang}
\author[1]{Shengxin Lu}
\author[1]{Lijun Wu}
\author[1]{Bin Wang}
\author[1]{Conghui He}

\affiliation[1]{Shanghai Artificial Intelligence Laboratory}
\affiliation[2]{King’s College London}
\affiliation[3]{East China University of Science and Technology}
\affiliation[4]{East China Normal University}
\affiliation[5]{Tongji University}
\affiliation[6]{Peking University}
\affiliation[7]{Fudan University}
\affiliation[8]{University of Science and Technology of China}

\abstract{
Optical Chemical Structure Recognition (OCSR) aims to translate molecular diagrams in scientific literature into machine-readable formats, but current systems remain unreliable on real-world images due to substantial visual and chemical complexity. We introduce \textbf{MOSAIC}, a dual-dimensional difficulty framework with 37 fine-grained labels that jointly characterize visual interference and chemical semantic challenges in molecular diagrams. Based on this framework, we construct \textbf{MolRecBench-Wild}, a benchmark of 5,029 structures from 820 recent chemistry papers, covering the full difficulty spectrum observed in real publications.
To enable faithful semantic evaluation beyond SMILES and MolFile, we propose \textbf{CARBON}, a representation language capable of expressing valence variations, icon-based groups, and other non-standard chemical semantics. We further adopt a dual-track evaluation protocol supporting both CARBON and SMILES outputs for broad model compatibility.
Comprehensive experiments over 18 OCSR-capable models reveal severe performance degradation on \textbf{MolRecBench-Wild}, exposing a large gap between previous patent benchmarks and real-world academic scenarios.
}

\date{\today}
\metadata[Equal contribution]{Haote Yang, Hui Wang, Chen Zhu, Jingchao Wang, Linye Li, Hongbin Lai, Huijie Ao, Yongxuan Lyu, Jiang Wu}
\metadata[Project leader]{Jiang Wu}
\correspondence{Conghui He, \email{heconghui@pjlab.org.cn}}

\metadata[Code]{\url{https://github.com/opendatalab/MolRecBench-Wild}}
\metadata[Dataset]{\url{https://huggingface.co/datasets/opendatalab/MolRecBench-Wild}}

\begin{document}

\maketitle

\section{Introduction}
\label{sec:intro}

\begin{figure}[t]
  \centering
  \includegraphics[width=0.7\linewidth]{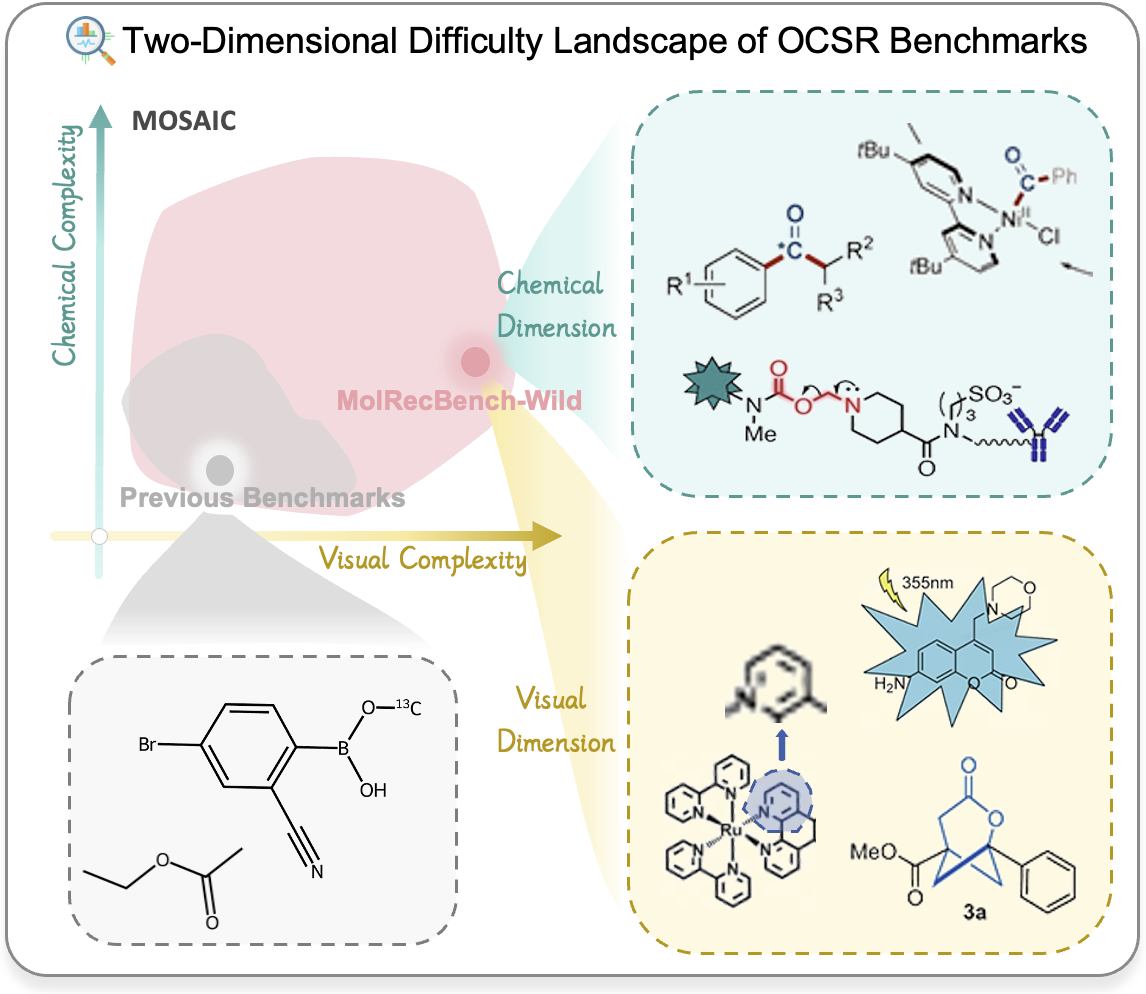}
  \caption{Our dual-dimensional difficulty landscape reveals a pronounced gap: existing OCSR benchmarks cluster in the simplest region of chemical and visual complexity, while MolRecBench-Wild, sourced from real literature images, extends deep into the high-difficulty domain in both visual and chemical dimensions.}
  \label{fig:why_molrecbench_wild}
\end{figure}

Optical Chemical Structure Recognition (OCSR) aims to convert molecular structure diagrams in chemical literature into machine-readable formats such as SMILES~\cite{weininger1988smiles}, graph structures, or MolFiles. It is a foundational technology for building large-scale, high-quality AI for Chemistry datasets and advancing the field. After years of development, OCSR has made significant progress, with current models performing excellently on mainstream evaluation benchmarks. For instance, MolGrapher~\cite{morin2023molgrapher} and MolParser~\cite{fang2025molparser} both achieve over 90\% Exact Match accuracy on benchmarks like USPTO~\cite{qian2023molscribe} and UOB\footnote{\url{http://www.cs.bham.ac.uk/research/groupings/reasoning/sdag/chemical.php}}. However, as shown in Figure~\ref{fig:why_molrecbench_wild} and Table~\ref{tab:dataset}, these benchmarks, although derived from real papers or patents, still consist of overly simple and homogeneous molecular diagrams, which differ greatly from the complexity and diversity of molecular structures found in recent mainstream chemistry journals. Existing models perform far below expectations in real-world challenging scenarios~\cite{rajan2025marcus}, suggesting that current public benchmarks can no longer support breakthroughs in molecular recognition technology. There is an urgent need to develop new benchmarks that are more representative of real-world applications, offering both greater difficulty and diversity.

\begin{figure*}[t]
  \centering
  \includegraphics[width=1.0\linewidth]{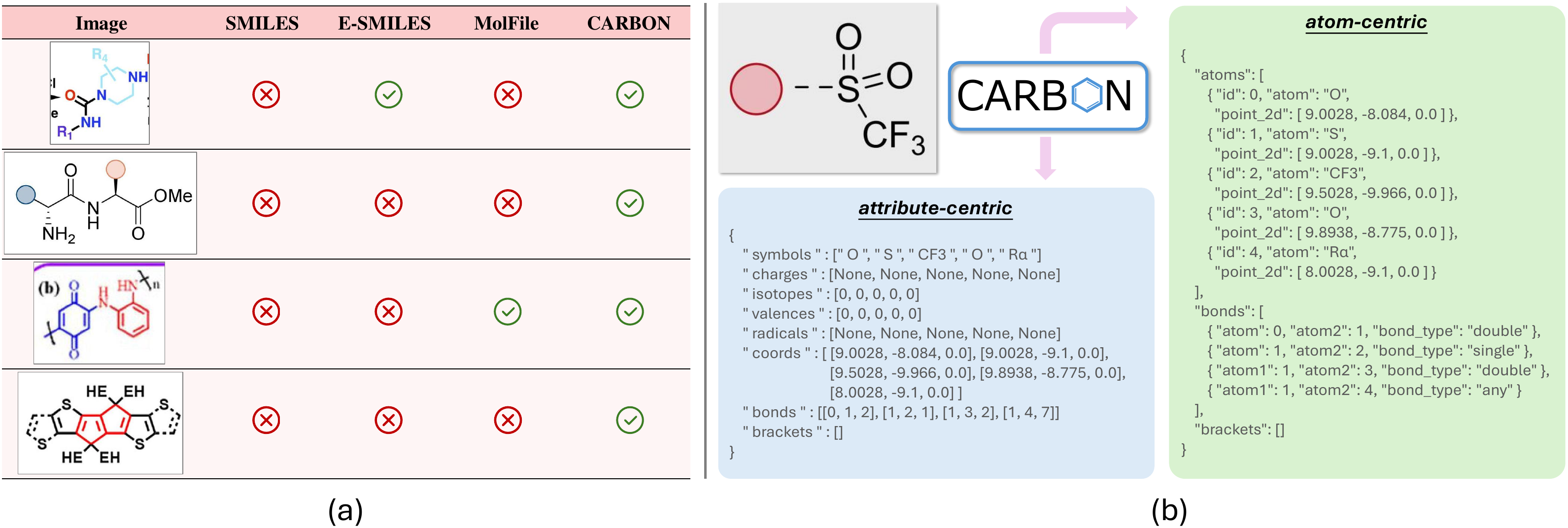}
  \caption{Sub-figure (a) illustrates the coverage of existing molecular representation methods. Current approaches fail to handle more complex cases, such as icon-based Markush structures, non-standard bonds, and other atypical patterns. Sub-figure (b) presents the CARBON representation of a given molecular example, which provides two complementary formats tailored for different use cases.}
  \label{fig:representation}
\end{figure*}

Based on an extensive analysis of real literature, we propose the \textbf{M}olecular \textbf{O}ptical-\textbf{S}emantic \textbf{A}ssessment of \textbf{I}mage \textbf{C}omplexity (\textbf{MOSAIC}), a dual-dimensional difficulty assessment framework specifically designed for molecular structure recognition tasks. As shown in Figure~\ref{fig:why_molrecbench_wild}, \textbf{MOSAIC} includes two orthogonal dimensions: visual representation and chemical semantics, defining 37 fine-grained difficulty labels. This framework provides a comprehensive, scientific, accurate, and objective way to assess the difficulty of specific samples in OCSR tasks. 

Based on \textbf{MOSAIC}, we developed \textbf{MolRecBench-Wild}, a molecular structure recognition benchmark derived from real chemical papers, which balances visual interference and complex chemical structures to provide a more challenging evaluation platform for assessing the robustness and generalization of models in real-world scenarios. The benchmark includes 5029 molecular structure diagrams from 820 chemistry papers, covering all labels in both the visual representation and chemical semantics dimensions. Of these, 93.29\% of samples have at least one \textbf{MOSAIC} difficulty label, and 42\% have labels in both dimensions. As shown in Table~\ref{tab:dataset}, compared to existing benchmarks, \textbf{MolRecBench-Wild} more accurately reflects the actual difficulty and challenges of molecular structure diagrams in current chemical literature, offering a more precise evaluation of OCSR models and general VLM performance.

Due to the complexity of chemical semantics in real-world molecular structure diagrams, existing representations like SMILES~\cite{weininger1988smiles}, E-SMILES~\cite{fang2025molparser}, and MolFile cannot fully capture all information, as described in~\cite{krasnov2024comparing} (Figure~\ref{fig:representation} (a)). To address this, we propose the \textbf{C}omplex \textbf{A}tomic \textbf{R}epresentation and \textbf{B}onding \textbf{O}bject \textbf{N}otation (\textbf{CARBON}), a molecular representation method designed for molecular recognition tasks (Figure~\ref{fig:representation} (b)). \textbf{CARBON} accurately expresses the true chemical state of complex molecular structures. Based on the \textbf{CARBON} annotation system, we can thoroughly assess OCSR models' ability to understand and parse all chemical semantic information in molecular diagrams. Additionally, to accommodate existing OCSR models that use SMILES output, we also evaluated these models and their inference methods on samples from MolRecBench-Wild with available SMILES representations. This dual-track evaluation mechanism ensures comparability with existing technologies while precisely pinpointing the model’s performance boundaries in real, complex scenarios.

We systematically evaluated five categories and eighteen OCSR-capable models on \textbf{MolRecBench-Wild}. 
The consistently low accuracy across SMILES, Simplified Graph, and Graph outputs highlights the substantial difficulty of recognizing molecular structures in real-world depictions. These results demonstrate the need for more realistic evaluation settings and reveal a critical gap between current OCSR solutions and practical application demands.
The contributions of this paper are as follows:

\begin{enumerate}
    \item We built MOSAIC, the first difficulty assessment system for molecular structure recognition tasks and developed MolRecBench-Wild based on this system. This benchmark is sourced from real chemical papers and reflects the complexity and diversity of both visual representation and chemical semantic dimensions, closely matching real-world application scenarios.
    \item We proposed CARBON, a molecular representation method that can express complex molecular information, providing effective support for constructing MolRecBench-Wild’s ground truth and ensuring accurate evaluations.
    \item We systematically evaluated the performance of 18 models in real-world scenarios, revealing the robustness flaws of existing methods and providing directional guidance for the development of molecular recognition tasks.
\end{enumerate}

\section{Related Work}

\subsection{Molecular Representation Methods}
Existing molecular representations mainly fall into two categories: string-based and rule-based. The former, represented by \textbf{SMILES}~\cite{weininger1988smiles}, is compact and efficient but struggles to describe complex structures such as metal coordination, stereochemistry, valence states, and other intricate details. Extensions like \textbf{CXSMILES}~\cite{ChemAxon_CXSMILES_2025} add stereochemical and charge information but lack consistent syntax; E-SMILES~\cite{fang2025molparser} improves on SMILES to support more complex structures, yet still doesn't address all its shortcomings; \textbf{SELFIES}~\cite{Krenn2020SELFIES} ensures syntactic validity but sacrifices expression efficiency and readability. On the other hand, rule-based formats like \textbf{MolFile} can explicitly record atomic and bond information, but they are verbose and difficult to adapt to non-standard bonds and structures.

In OCSR tasks, these representations often face limitations in expressiveness and structural redundancy, making it hard to cover the complexity of molecules in chemical literature. To address this, we propose \textbf{CARBON}, a semantically structured and syntactically extensible molecular representation language. \textbf{CARBON} can accurately represent coordination, mixed valence states, and complex bond types while maintaining compatibility with traditional graph structures, offering a more unified, compact, and chemically complete output format for OCSR models. Refer to Figure~\ref{fig:representation} for the comparison between different molecule representation methods.

\subsection{OCSR Task}
Early OCSR systems, such as \textbf{OSRA} \footnote{\url{https://sourceforge.net/p/osra/wiki/Validation}}, relied on rule-based matching and template retrieval for analyzing chemical structure images. However, these approaches were sensitive to image noise, resolution, and drawing style variations, making them unsuitable for complex scenarios like scanned or hand-drawn structures.

With the rise of deep learning, two main approaches for OCSR tasks have emerged. The first is the direct prediction of SMILES. \textbf{DECIMER}~\cite{rajan2020decimer} introduced the Transformer architecture, enabling end-to-end translation from chemical structure images to SMILES, marking a new phase of neural network-based OCSR. \textbf{SwinOCSR}~\cite{xu2022swinocsr} used Swin Transformer as a visual encoding backbone for end-to-end conversion. \textbf{MPOCSR}~\cite{lin2024mpocsr} combined convolutional networks and Transformers to jointly model multi-scale features, improving the capture of both local and global chemical semantics in images. Recently, \textbf{MolParser}~\cite{fang2025molparser} continued DECIMER's approach but adopted the E-SMILES prediction format, allowing it to support more Makuṣh-style molecules. Meanwhile, \textbf{OCSU}~\cite{fan2025ocsu} treated SMILES prediction as a pre-training task for other chemical reasoning tasks, offering a new approach to multi-task learning.

The second approach involves predicting the molecular graph first and then using it to predict SMILES. \textbf{MolGrapher}~\cite{morin2023molgrapher} explicitly incorporates atomic position information to reconstruct the molecular graph and generate SMILES, enhancing the utilization of graph structure information. \textbf{MolScribe}~\cite{qian2023molscribe} improved this by using a Transformer decoder to directly predict the full molecular graph, significantly improving structure reconstruction accuracy. \textbf{MolNexTR}~\cite{chen2024molnextr} combined MolScribe's graph generation with MPOCSR's multi-scale visual modeling, enhancing robustness and generalization. \textbf{GTR-Mol-VLM}~\cite{wang2025gtr} applied visual-language models (VLMs) to OCSR tasks, introducing a graph traversal-based molecular structure prediction framework and new evaluation metrics.

In addition to expert models, general multimodal large models, such as \textbf{GPT-4o}~\cite{openai2024gpt4o}, \textbf{Qwen-VL-Max}~\cite{qwen2025qwen25vl}, \textbf{GLM-4.5V}~\cite{glm2025glm45v}, \textbf{Gemini 2.5 Pro}~\cite{google2025gemini25}, and \textbf{InternVL3.5}~\cite{internvl2025internvl35}, also include OCSR-related data in their training sets. Several works (e.g., \textbf{ChemVLM}~\cite{Li2025ChemVLM}, \textbf{ChemDFM-X}~\cite{Zhao2024ChemDFMX}, \textbf{ChemMLLM}~\cite{Tan2025ChemMLLM}) have fine-tuned these multimodal models in the chemistry domain, often including OCSR sub-tasks. However, compared to expert models specifically designed for this task, these general-purpose or domain-specific large models still show a noticeable performance gap in recognition.

Although the above methods have achieved excellent results on public benchmarks, recent research~\cite{krasnov2024comparing} shows that when models are applied to real-world chemical literature images with significant noise and inconsistent quality, their recognition performance is still significantly lower than on standard test sets. This highlights the key challenges OCSR models face in terms of generalization and adaptation to real-world data.

\begin{figure*}[t]
  \centering
  \includegraphics[width=1.0\linewidth]{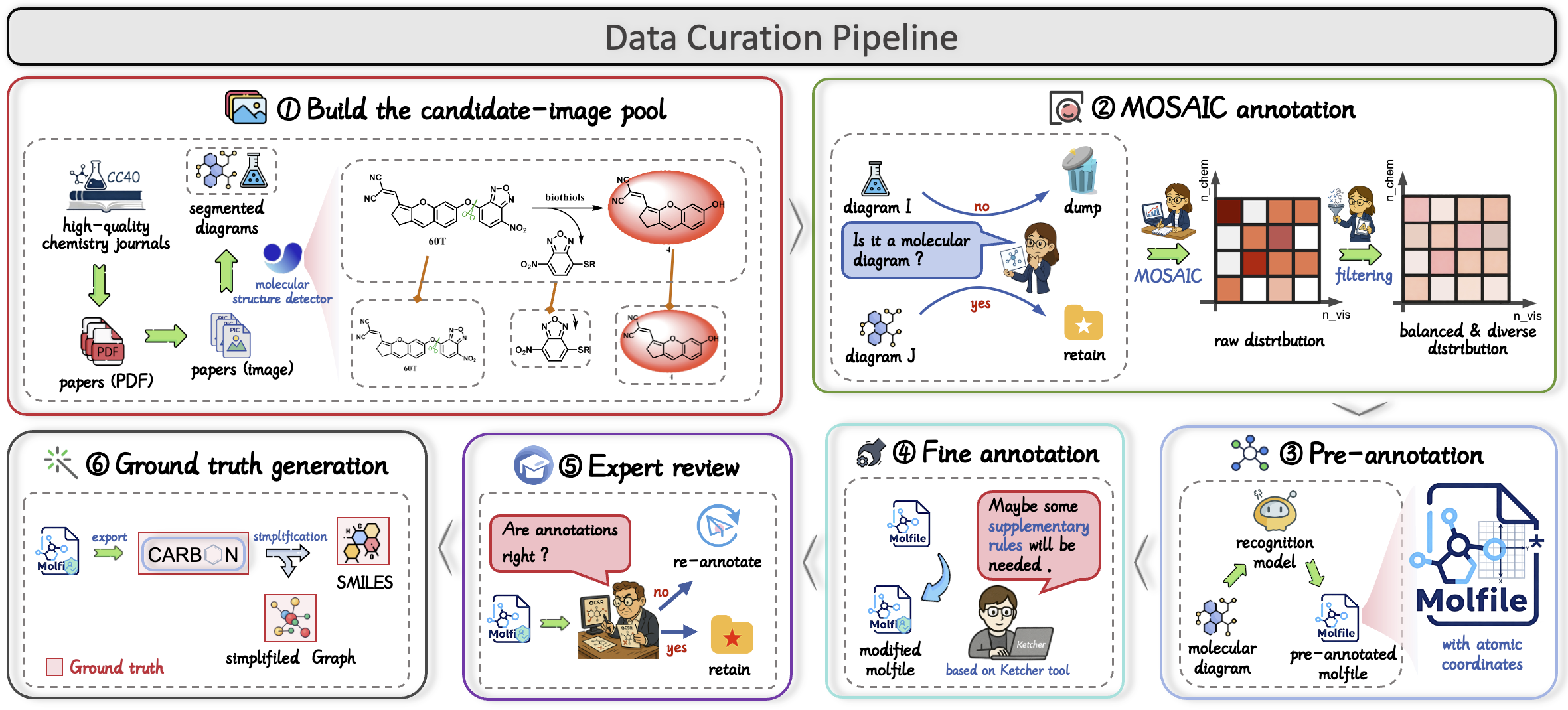}
  \caption{An overview of the \textbf{MolRecBench-Wild} data curation pipeline. All molecules are sourced from leading chemical journals published in recent years. All annotations are produced and/or verified by domain experts, ensuring high-quality and reliable labeling. The ground truth is eventually saved in CARBON, simplified Graph, and SMILES formats.}
  \label{fig:data_curation}
\end{figure*}

\subsection{Existing OCSR Benchmarks}
Existing OCSR benchmarks, such as the \textbf{DECIMER} Dataset~\cite{DECIMERV2_Zenodo_JPO}, \textbf{CLEF}~\cite{Sadawi2012MolRecCLEF}, \textbf{Staker}~\cite{Staker2018DeepOCSR}, \textbf{USPTO}~\cite{qian2023molscribe}, \textbf{USPTO-30K}~\cite{morin2023molgrapher}, \textbf{JPO}~\cite{DECIMERV2_Zenodo_JPO}, and \textbf{MolMole Patents}~\cite{MolMolePatent300_2025}, are mostly sourced from patents or automatically generated images, featuring standardized drawings, clean backgrounds, high resolution, and minimal noise. These datasets have limited visual and chemical complexity: visually, they lack scanning artifacts, print noise, text or layout interference; chemically, they mainly contain regular molecular structures and rarely cover real-world scenarios like multi-group polymers, non-standard bonds, mixed valence states, free radicals, and difficult-to-symbolize elements. Even when such cases appear, they are often not reflected in the ground truth annotations. While WildMol attempts to address these issues by collecting more complex samples and improving SMILES representation, it has made only limited progress, with many hard cases still uncovered and the expressive capacity of E-SMILES remaining limited.

To address these gaps, \textbf{MolRecBench-Wild} builds a dataset from real chemical paper images, balancing visual interference with complex chemical structures, providing a more challenging evaluation platform for testing model robustness and generalization in real-world scenarios.
\section{MolRecBench-Wild}
\subsection{Molecular Optical-Semantic Assessment of Image Complexity (MOSAIC)}
Existing molecular structure recognition evaluation benchmarks consist of overly simple and homogeneous samples, which differ significantly from the complex molecular structures found in recent mainstream chemistry journals. To develop an evaluation benchmark that accurately reflects the diversity and difficulty of molecular structure diagrams in real-world scenarios, it is essential to objectively and scientifically measure the difficulty of specific samples in the OCSR task. To this end, we propose the \textbf{M}olecular \textbf{O}ptical-\textbf{S}emantic \textbf{A}ssessment of \textbf{I}mage \textbf{C}omplexity (\textbf{MOSAIC}), a difficulty assessment framework specifically designed for molecular structure recognition.

Based on an analysis of numerous real literature sources, we define \textbf{MOSAIC} with two orthogonal dimensions: visual presentation and chemical semantics. In the visual presentation dimension, we focus on how the style of molecular structure diagrams affects recognition. As shown in Figure~\ref{fig:why_molrecbench_wild}, current benchmarks primarily feature diagrams with clean backgrounds, high resolution, and minimal noise. In contrast, real literature often includes structures with background colors, blurriness, arrows, text interference, and background images that hinder model perception. In the chemical semantics dimension, we focus on the complexity of the chemical information conveyed by the diagrams. As shown in Figure~\ref{fig:why_molrecbench_wild}, existing benchmarks only contain regular structures that can be represented with standard SMILES syntax. Real-world molecular diagrams, however, often involve metal coordination, oxidation state changes, and multi-center reactions, with some chemical bonds that cannot be represented in SMILES or MolFile formats. Some R groups use graphical symbols such as circles, which are difficult to be processed by the tokenizer of the sequence model.

Based on these two dimensions, we further systematize and formally define the various difficulty aspects of molecular structure diagrams, resulting in 18 visual presentation labels and 19 chemical semantics labels, as detailed in Appendix~\ref{apx:b}. For each molecular diagram, we manually annotate whether it meets the definition criteria for each label, enabling a fine-grained characterization of sample recognition complexity and diversity.

\begin{table}[t]
\caption{Comparison of major open-source OCSR benchmarks containing more than 500 molecular images. \textbf{Source} denotes whether the dataset originates from published papers or patent documents. \textbf{VDL} and \textbf{CDL} denote the numbers of Visual Difficulty Labels and Chemical Difficulty Labels, respectively, according to the \textbf{MOSAIC} framework. \textbf{DA} indicates whether Difficulty Annotation is available. \textbf{Source Info} specifies whether each molecular image includes explicit provenance information, such as the patent ID, literature source, or figure reference.}
\centering
\resizebox{\textwidth}{!}{
\small
\setlength{\tabcolsep}{5pt}  %
\begin{tabular}{c c c c c c c c c}
\toprule
Benchmark & \# Image & Source & Image Size (mean $\pm$ std) &
Ground Truth & \# VDL & \# CDL & DA & Source Info\\ 
\midrule
Staker & 49,557 & Patent &
$256\pm0 \times 256\pm0$ & SMILES & $\textless 10$ & $\textless 3$ & $\times$ & $\checkmark$ \\
UOB & 5,740 & Patent &
$762\pm73 \times 412\pm125$ & MolFile & $\textless10$ & $\textless3$ & $\times$ & $\times$ \\
USPTO & 5,638 & Patent &
$688\pm234 \times 437\pm178$ & MolFile & $\textless10$ & $\textless3$ & $\times$ & $\checkmark$ \\
CLEF & 906 & Patent &
$634\pm193 \times 385\pm135$ & MolFile & $\textless10$ & $\textless3$ & $\times$ & $\checkmark$ \\
USPTO-30K abbreviated & 9,998 & Patent &
$660\pm253 \times 411\pm166$ & MolFile & $\textless10$ & $\textless3$ & $\times$ & $\checkmark$ \\
USPTO-30K clean & 10,000 & Patent &
$675\pm249 \times 437\pm171$ & MolFile & $\textless10$ & $\textless3$ & $\times$ & $\checkmark$ \\
USPTO-30K large & 10,000 & Patent &
$1296\pm418 \times 857\pm295$ & MolFile & $\textless10$ & $\textless3$ & $\times$ & $\checkmark$ \\
MolMole Patents & 2,469 & Patent &
$322\pm181 \times 202\pm96$ & MolFile & $\textless10$ & $\textless3$ & $\times$ & $\checkmark$ \\
MolRecBench-Wild (ours) & 5,064 & Article &
$527\pm253 \times 378\pm175$ & CARBON & 18 & 19 & $\checkmark$ & $\checkmark$ \\
\bottomrule
\end{tabular}

}
\label{tab:dataset}
\end{table}



We define the \textbf{MOSAIC} metric as a pair (${N_{vis}, N_{chem}}$), where $N_{vis}$ represents the number of labels in the visual presentation dimension classified as positive for the sample, and $N_{chem}$ represents the number of labels in the chemical semantics dimension classified as positive. This metric quantifies the difficulty level of molecular structure diagrams in both the visual presentation and chemical semantics dimensions. Based on this, for any molecular structure diagram dataset, the distribution of samples in the ($N_{vis}, N_{chem}$) two-dimensional space can be calculated, and a distribution matrix can be constructed, each cell in the matrix represents the number of molecular structure diagram samples corresponding to that coordinate position. (Appendix~\ref{apx:c}).

\subsection{Data Curation}

As shown in Figure~\ref{fig:data_curation}, the data collection and annotation process for \textbf{MolRecBench-Wild} consists of six stages:

1. \textbf{Building the Candidate Image Pool}: We collected approximately 820 CC-BY-4.0 licensed papers from 7 high-level chemistry journals (list in Appendix~\ref{apx:f}). After converting them to image format, we used DocLayout-YOLO and our internal YOLO-based molecular structure detector to extract bounding boxes for all molecular diagrams. After cropping, we obtained 7000 molecular structure images, forming the \textbf{MolRecBench-Wild} candidate pool.

2. \textbf{Initial Screening, \textbf{MOSAIC} Annotation, and Refinement}: We first manually performed an initial screening, selecting around 6548 molecular structure images from the candidate pool. Then, we manually annotated them with \textbf{MOSAIC} labels. Based on these labels, we further selected 5029 samples to ensure an even distribution across the ($N_{vis}, N_{chem}$) two-dimensional space, covering a wide range of challenging samples.

3. \textbf{Molecular Structure Pre-annotation}: We used an internal molecular recognition model to predict the molecular structures in the selected diagrams, generating pre-annotations for later correction. Unlike standard SMILES output, this model predicts the exact positions of atoms in the molecular diagram and outputs the molecular information in graph format, with predictions fully aligned with the original image layout, facilitating subsequent manual review and correction.

4. \textbf{Manual Refinement}: Using the Ketcher tool, we manually checked and corrected the model’s predicted molecular diagrams, following detailed annotation guidelines to fully cover all \textbf{MOSAIC} ``chemical semantics'' labels. For chemical semantics not supported by the standard MolFile format, we created supplementary rules to ensure the final ``modified'' MolFile accurately and completely records these semantic details.

5. \textbf{Quality Control}: All annotated samples undergo at least two rounds of strict quality checks. Samples that do not meet the standards are returned for re-annotation until they meet the required quality.

6. \textbf{Ground Truth Generation}: All ground truth of annotated samples are then transformed into \textbf{CARBON} and SMILES format (if possible).

To support large-scale collaborative annotation, we developed an integrated online annotation platform that includes task distribution, annotation tools, quality checks, rework management, and progress monitoring, ensuring efficiency, control, and traceability throughout the annotation process. We recruited 47 annotators with chemistry backgrounds for the project (detailed in Appendix~\ref{apx:e}), with a total of 152 person-days spent on the entire annotation process.

\subsection{Dataset Statistics}

Table~\ref{tab:dataset} compares the characteristics of the MolRecBench-Wild dataset, built in this study, with existing OCSR datasets across multiple dimensions. \textbf{MolRecBench-Wild} covers 19 categories of chemical difficulty and 18 categories of visual difficulty, highlighting its advantages in chemical structure diversity and image complexity. Unlike other datasets that typically use SMILES or Mol file annotations, \textbf{MolRecBench-Wild} uses high-quality annotations of \textbf{CARBON} format, which more intuitively express complex chemical structure information. Additionally, this dataset systematically annotates molecular difficulty levels and sources, significantly improving the interpretability, reliability, and traceability, providing a more rigorous foundation for training and evaluating OCSR models.

\begin{figure}
    \centering
    \includegraphics[width=0.7\linewidth]{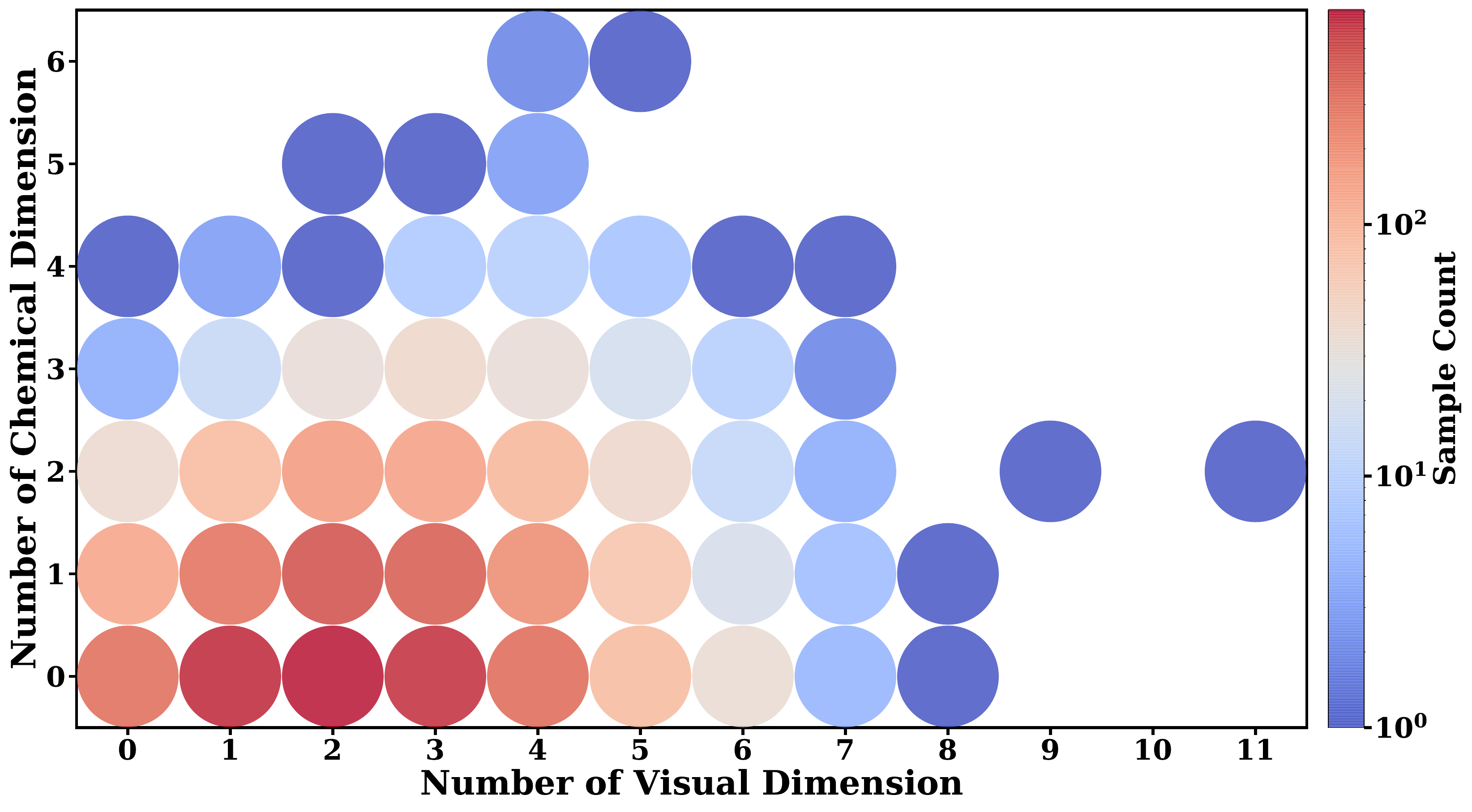}
    \caption{The figure illustrates the joint distribution of molecular images in the MolRecBench-Wild dataset across the number of visual dimensions and chemical dimensions. Each circular bubble represents a unique combination of these two dimensions, with the color intensity indicating the number of samples. The results show that samples with lower-dimension combinations (bottom-left region) are more concentrated, whereas those with higher-dimension combinations (upper-right region) are relatively sparse.}
    \label{fig:distribution}
\end{figure}

Figure~\ref{fig:distribution} presents the two-dimensional distribution heatmap of molecular image difficulty labels, showing that the molecules in \textbf{MolRecBench-Wild} generally possess a rich set of difficulty labels in both visual presentation and chemical semantics dimensions, with a relatively balanced distribution of different difficulty labels. This indicates a reasonable design of the task difficulty distribution in the dataset. The statistical results of the difficulty label types for \textbf{MolRecBench-Wild} molecular images are detailed in Appendix\ref{apx:c}.

\section{Complex Atomic Representation and Bonding Object Notation (CARBON)}
To address the limitations of existing molecular representations in expressing complex chemical structures, we propose a new molecular description method called \textbf{C}omplex \textbf{A}tomic \textbf{R}epresentation and \textbf{B}onding \textbf{O}bject \textbf{N}otation (\textbf{CARBON}), specifically designed for molecular recognition tasks. As shown in Figure~\ref{fig:representation}, \textbf{CARBON} consists of two complementary forms: atom-centric and attribute-centric. The former is suited for model training and inference, while the latter is better for data statistical analysis. These two forms can be converted to meet different application needs.

\textbf{CARBON} offers enhanced expressive capabilities: (1) It supports \textbf{non-standard bond types}, such as dashed dative bonds, bold bonds, bold double bonds, and dashed double bonds, which are commonly found in literature images but unsupported by MolFile. (2) It supports \textbf{a wide range of atomic attributes}, including oxidation states, radicals, isotopes, and non-integer charges. (3) It supports \textbf{repetitive structures}, allowing explicit representation of multi-groups and polymers. (4) It supports \textbf{image-level coordinate information to preserve spatial distribution features of structure diagrams}, facilitating visual molecular recognition tasks.

\textbf{CARBON} has three core features: (i) \textbf{Extensibility}: Additional fields can be added to support more atomic properties and bond types, such as expressing non-central bonds and overall molecular charges through the “atom group” concept. (ii) \textbf{Flexibility}: The two forms are designed for different scenarios, can be converted into each other, and, under certain conditions, can be simplified into SMILES format. (iii) \textbf{Conciseness and Readability}: Compared to MolFile, CARBON structures are more compact and semantically clearer, making them especially suitable for molecular recognition and modeling tasks.
\section{Evaluation Protocol}

\begin{figure}[t]
    \centering
    \includegraphics[width=1\linewidth]{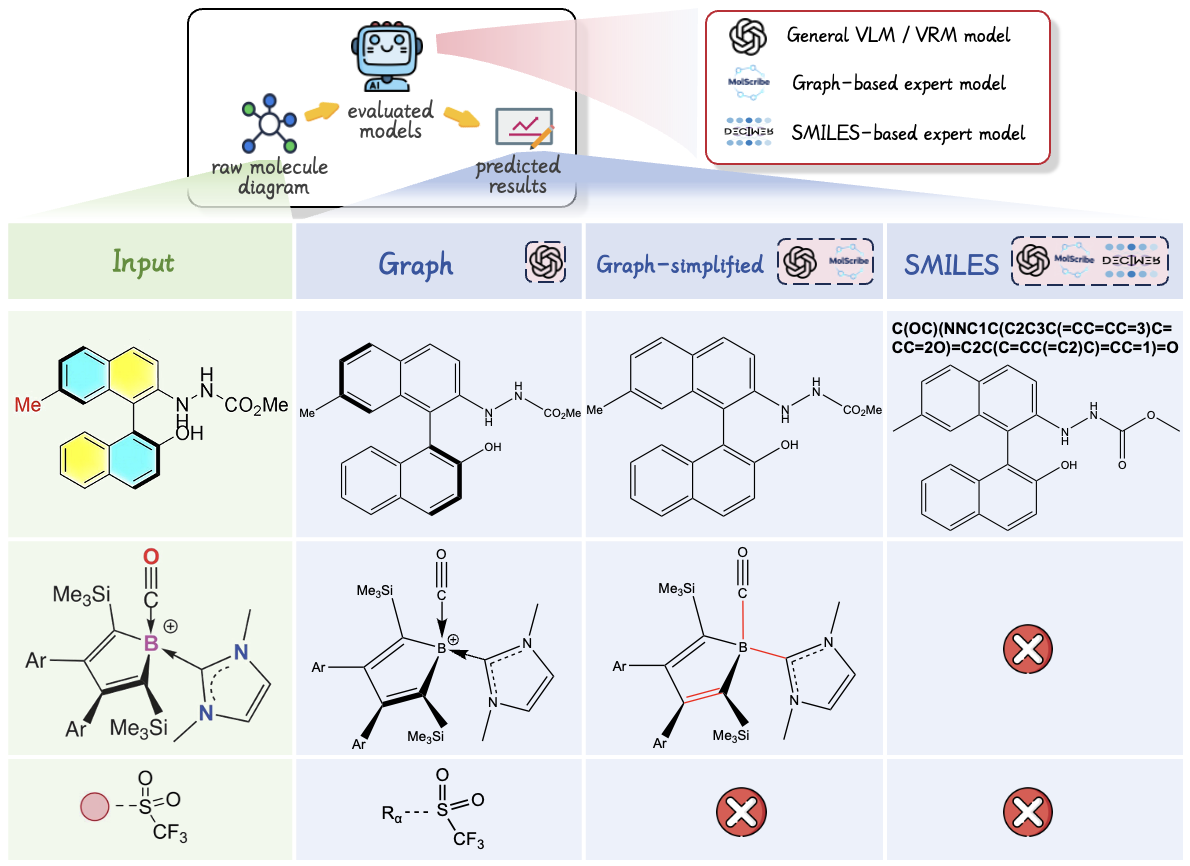}
    \caption{Comparison of model outputs on molecules spanning different levels of structural complexity. For each input molecule diagram, we display the predicted results in Graph, Simplified Graph, and SMILES formats. Models with different capabilities can be evaluated under the corresponding evaluation protocols.}
    \label{fig:evaluation}
\end{figure}

To comprehensively assess the model's understanding and reconstruction ability of molecular structures, this study proposes a unified evaluation protocol that enables comparability across different types of models. For a given molecular image (Figure~\ref{fig:evaluation}), the model outputs either a SMILES sequence or a molecular graph (Graph) based on its characteristics. Since this work involves general visual-language models (VLM/VRM), SMILES expert models, and Graph expert models, we have designed two evaluation methods—based on Graph and based on SMILES—within the evaluation process.

\textbf{Graph-based Evaluation}: For vision language models (VLMs) with strong instruction-following capabilities, new chemical bond types in the dataset can be recognized through prompt design, without the need for additional training. In contrast, some expert models can only recognize bond types that were present during the training process. Thus, we divide the Graph-based evaluation into two categories: Graph and Simplified Graph. In the Graph evaluation, we perform precise matching based on atom types, chemical bonds, and parentheses information; while in Graph-simplified, to reduce task difficulty, only atom symbols and common chemical bonds are matched, with complex bond types simplified to basic bond types.

\textbf{SMILES-based Evaluation}: Some expert models can only output SMILES sequences and cannot predict the atomic spatial positions or topological structure of the molecule. Therefore, we have designed a SMILES-based evaluation method. It should be noted that due to the diverse chemical bonds in the dataset, which cannot be fully represented in standard SMILES, the bonds in the SMILES labels in this study are simplified to ensure consistency and operability in the evaluation.
\section{Experiments}
\begin{table}[h]
\caption{Evaluation results of different methods on MolRecBench-Wild. Underlined values indicate the best results within each class, and bolded values represent the overall best results across all classes. The version of InternVL3.5 is InternVL3.5-241B-A28B. Symbol \textsuperscript{†} means the model is finetuned on chemical tasks.}
\centering
\begin{tabular}{lcccc}
\Xhline{2pt}
Method & SMILES  & \makecell{Simplified \\ Graph}  & Graph \\ 
\Xhline{1.5pt}
\multicolumn{4}{c}{SMILES-based Expert Models} \\
\Xhline{0.5pt} 
OCSU                           & 6.06                        & -            & -     \\
DECIMERv2.2                   & $\underline{22.84}$                        & -            & -     \\[4pt]
\Xhline{1pt}
\multicolumn{4}{c}{Graph-based Expert Models} \\
\Xhline{0.5pt} 
MolGrapher                     & 20.33                        & 22.81        & -     \\
MolNexTR                       & 40.90                        & 34.42        & -     \\
MolScribe                      & $\underline{\textbf{41.05}}$                        & 34.47        & -     \\
GTR-Mol-VLM                    & 40.43 & $\underline{\textbf{35.22}}$        & -     \\[4pt]
\Xhline{1pt}
\multicolumn{4}{c}{Vision Language Models} \\
\Xhline{0.5pt} 
GPT-4o                         & 7.94                         & 3.74         & 2.94  \\
Qwen-VL-Max                    & 6.95                        & 5.83         & $\underline{3.66}$  \\
InternVL3.5                    & $\underline{25.60}$          & $\underline{6.88}$         & 3.08  \\
ChemVLM\textsuperscript{†}     & 4.79                         & -            & -     \\
ChemDFM-X\textsuperscript{†}   & 9.75                         & -            & -     \\[4pt]
\Xhline{1pt}
\multicolumn{4}{c}{Vision Reasoning Models} \\
\Xhline{0.5pt} 
GPT-5                         & 19.68                        & 10.00         & 8.19   \\
Seed1.6-Thinking              & 15.60                        & 7.14         & 4.61  \\
Intern-S1                       & 18.98                        & 6.62         & 3.46  \\
Gemini 2.5 Pro                 & $\underline{30.06}$          & $\underline{15.67}$        & $\underline{\textbf{13.04}}$ \\
GLM-4.5V                       & 12.13                        & 7.89         & 4.26  \\[4pt]
\Xhline{1pt}
\multicolumn{4}{c}{Tools} \\
\Xhline{0.5pt} 
Mathpix                       & $\underline{27.88}$           & -            & -     \\
Logics-Parsing                 & 15.47                        & -            & -     \\ 
\Xhline{2pt}
\end{tabular}

\label{tab:result}
\end{table}

\subsection{Settings}
To comprehensively evaluate model performance on real-world chemical structure recognition, we benchmarked a diverse set of representative systems:
Specialized OCSR models: e.g., \textbf{OCSU}~\cite{fan2025ocsu} and \textbf{DECIMER v2.2}~\cite{rajan2020decimer}, which generate SMILES strings from chemical images.
Graph-based models: including \textbf{MolGrapher}~\cite{morin2023molgrapher}, \textbf{MolNexTR}~\cite{chen2024molnextr}, \textbf{MolScribe}~\cite{qian2023molscribe}, and \textbf{GTR-Mol-VLM}~\cite{wang2025gtr}, which directly infer molecular graph structures.
General-purpose multimodal large models: such as \textbf{GPT-4o}~\cite{openai2024gpt4o}, \textbf{GPT-5 }\footnote{\url{https://openai.com/index/introducing-gpt-5/}}, \textbf{Gemini 2.5 Pro}~\cite{google2025gemini25}, \textbf{GLM-4.5V}~\cite{glm2025glm45v}, \textbf{Seed1.6-Thinking} \footnote{\url{https://seed.bytedance.com/zh/blog/introduction-to-techniques-used-in-seed1-6}}, \textbf{Qwen-VL-Max}~\cite{qwen2025qwen25vl}, and \textbf{InternVL3.5-241B-A28B}~\cite{internvl2025internvl35}.
Domain-specific expert models: including \textbf{ChemVLM}~\cite{Li2025ChemVLM}, \textbf{ChemDFM-X}~\cite{Zhao2024ChemDFMX} tailored for chemical and AI4Sci model Intern-S1~\cite{bai2025intern}.
Document parsing tools: such as \textbf{Mathpix} \footnote{\url{https://mathpix.com/}}  and \textbf{Logics-Parsing}~\cite{chen2025logics}.
For ChemVLM, ChemDFM-X, and other domain expert models, we adopted the official prompt templates to match their original evaluation settings. For general-purpose VLMs, we applied a unified prompt template to ensure fair comparison. Full prompts are provided in Appendix~\ref{apx:d}.

\subsection{Main Results}
Tab.~\ref{tab:result} presents a comprehensive comparison of a diverse set of models on our proposed benchmark under three predicted formats—SMILES, Simplified Graph, and Graph—and several important observations emerge from the results. 

(1) All model categories achieve relatively low scores across formats, indicating that current mainstream approaches still struggle to robustly recognize molecular structures from visual inputs in realistic and noisy conditions. This highlights the substantial difficulty of our benchmark and reinforces its value in exposing the gap between controlled experimental settings and real-world OCSR challenges. 

(2) Across the SMILES and Simplified Graph evaluations, models such as GTR-Mol-VLM, MolScribe, and MolNexTR consistently outperform others, suggesting that graph-based architectures maintain stronger structural priors and thus exhibit better robustness when dealing with diverse molecular images. 

(3) Among VLMs (vision language models), Gemini 2.5 Pro demonstrates notably strong generalization ability, especially on the Graph metric, where it significantly surpasses other VLMs. This suggests that a vision reasoning capacity contributes meaningfully to complex molecular structure interpretation. 

(4) The VRMs (vision reasoning models) show a clear and substantial advantage over general-purpose VLMs and fine-tuned VLM variants, revealing that explicit reasoning mechanisms play a crucial role in enhancing model stability and accuracy in OCSR tasks. 

(5) Among open-source tools, Mathpix achieves the highest performance and exceeds Logics-Parsing by a large margin. 

Together, these observations outline not only the current landscape of OCSR model performance but also the promising directions for future research.

\begin{figure}[t]
    \centering
    \includegraphics[width=0.7\linewidth]{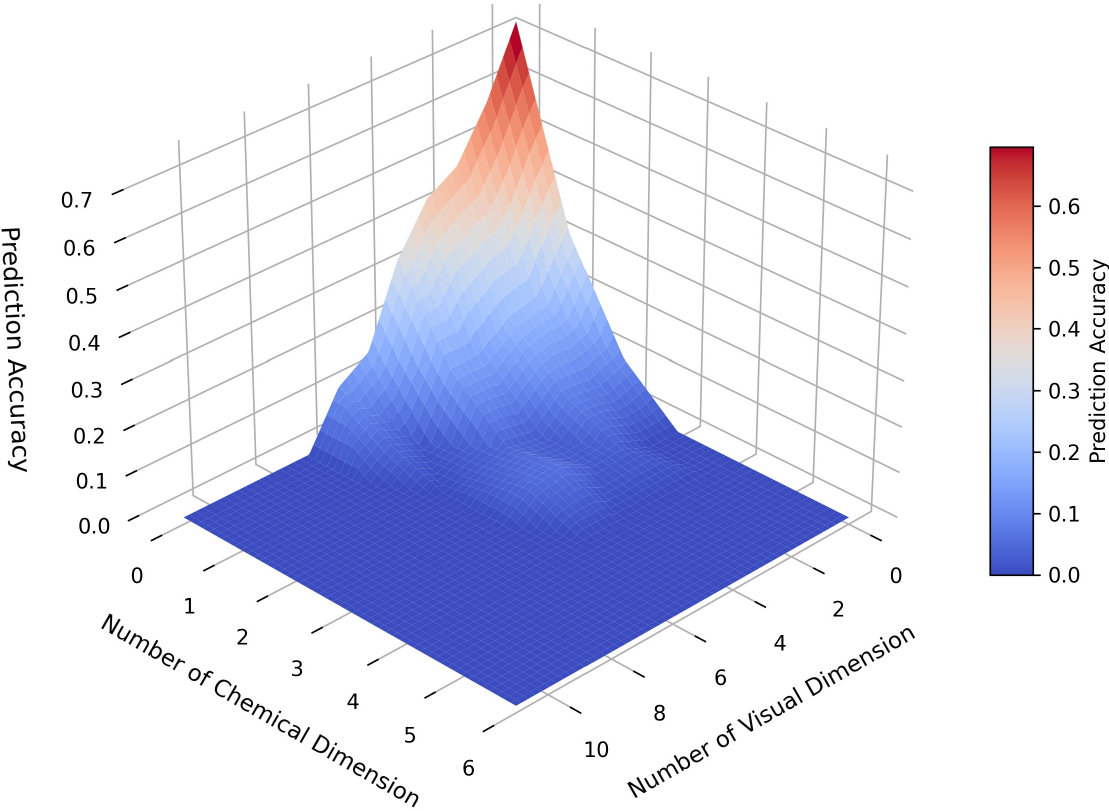}
    \caption{Accuracy of GTR-Mol-VLM under combinations of varying numbers of chemical and visual dimension challenges on Simplified Graph metric.}
    \label{fig:evaluation}
\end{figure}

\subsection{Analysis}

To validate the rationality of our designed MOSAIC, we visualized the accuracy of GTR-Mol-VLM under different combinations of visual and chemical challenges. As the number of challenges in either the visual or chemical dimension increases, the accuracy of GTR-Mol-VLM shows a declining trend.
This trend demonstrates that MOSAIC successfully constructs a controllable difficulty spectrum and can expose the sensitivity of current OCSR models to different types of perturbations. In particular, the sharp decline in accuracy under joint visual–chemical challenges highlights a key limitation of existing approaches and underscores the necessity of future models to achieve better robustness across heterogeneous perturbations.

Besides, across evaluated models, we observe a consistent performance hierarchy in which SMILES accuracy greater than Simplified Graph accuracy greater than Graph accuracy. 
This descending trend reflects the increasing difficulty of the evaluation protocols: SMILES prediction only requires the model to capture a canonicalized sequence representation, whereas Simplified Graph demands accurate reconstruction of the molecular scaffold with reduced atom-level detail. 
The full Graph representation poses the greatest challenge, requiring precise atom identities, bond orders, stereochemistry, and connectivity.
This monotonic degradation in performance demonstrates that our multi-protocol evaluation framework effectively captures different levels of structural granularity, providing a more comprehensive assessment of molecular recognition capabilities.

\section{Discussion}
This study demonstrates that existing OCSR methods still fall short of achieving reliable performance on real-world scientific literature images. The proposed two-dimensional difficulty framework provides a more fine-grained perspective for model evaluation and identifies key challenges for future model development. CARBON, as a novel molecular representation language, offers stronger semantic coverage and shows great potential to serve as a unified standard for complex molecular recognition tasks. Furthermore, we suggest that future OCSR research could advance along several directions, including vision–chemistry cross-modal pretraining, style-aware image augmentation and generalization, and unified structural language representation.
\section{Conclusion}
In this work, we reveal that current OCSR systems remain far from reliable in real-world chemical literature and attribute this gap to the lack of realistic difficulty modeling and semantically complete ground truth. We introduce MOSAIC, the first dual-dimensional difficulty framework for molecular diagrams, and use it to construct MolRecBench-Wild, a benchmark sourced entirely from recent chemical publications and covering the full range of visual and chemical complexities. We further propose CARBON, a representation language capable of expressing real-world chemical semantics beyond the limits of SMILES and MolFile, enabling precise and comprehensive evaluation. Experiments on 18 models show substantial performance degradation across output formats, highlighting the urgent need for more robust OCSR methods. By releasing our dataset, representation, and evaluation toolkit, we hope to establish a foundation for advancing reliable molecular structure recognition.


\clearpage
\newpage
\bibliographystyle{plainnat}
\setcitestyle{numbers}
\bibliography{paper}

\clearpage
\newpage

\beginappendix

\section{Case Study}
\label{apx:a}
To better analyze the performance of different methods under various evaluation metrics, we conducted a visual analysis of the results from the best-performing method among the five approaches.

\subsection{SMILES Qualitative Results}

\begin{figure}[h]
    \centering
    \includegraphics[width=1\linewidth]{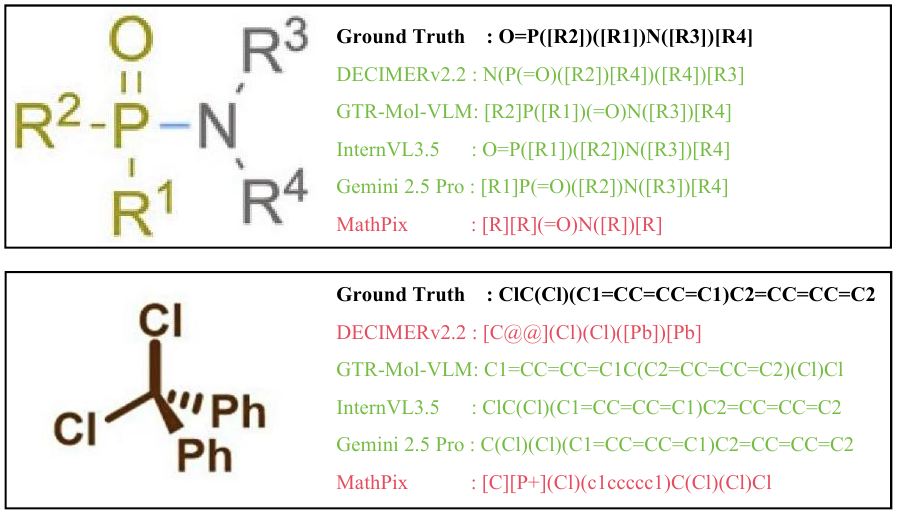}
    \caption{Comparison of SMILES results produced by different models.}
    \label{fig:smiles}
\end{figure}

As shown in \ref{fig:smiles}, we selected two samples for demonstration. 
In the first sample, although the final displayed SMILES are not absolutely identical, DECIMERv2.2, GTR-Mol-VLM, InternVL3.5, and Genmini 2.5 Pro all predicted the correct results while preserving the sequence numbers for different R values. On the contrary, MathPix not only represents all R groups in a unified format but also omits the prediction of P(phosphorus) atoms.
In the second sample, although the diagram depicts both solid and hollow wedge bonds, this molecule is actually achiral. GTR-Mol-VLM, InternVL3.5, and Gemini 2.5 Pro effectively filtered out these distractions and provided accurate predictions. However, DECIMERv2.2 misidentified the functional group '[Ph]' as '[Pb]' and introduced unnecessary chiral information. Additionally, MathPix also produced incorrect predictions.

\subsection{Simplified Graph Qualitative Results}

\begin{figure*}
    \centering
    \includegraphics[width=1\linewidth]{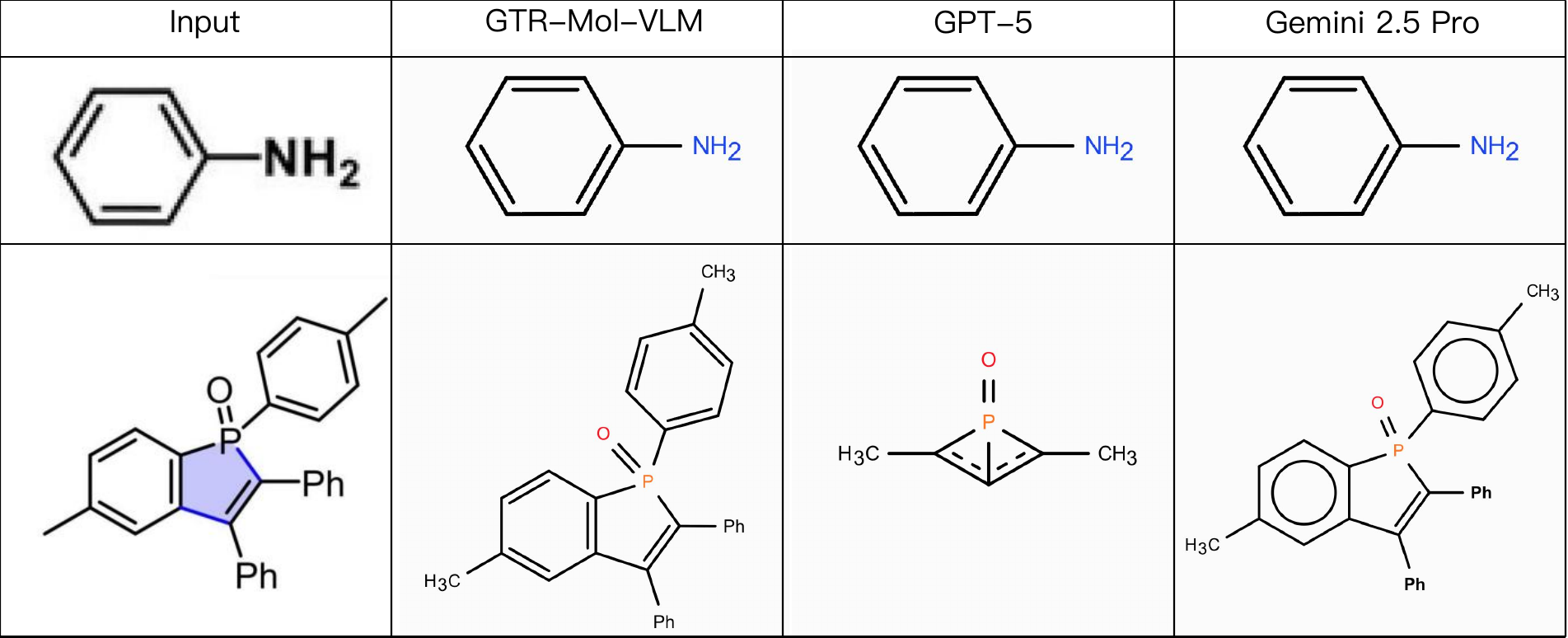}
    \caption{Comparison of Simplified Graph results produced by different models.}
    \label{fig:sgraph}
\end{figure*}

As shown in Fig. \ref{fig:sgraph}, we present the prediction results of the three methods in Simplified Graph. 
It is evident that for simple molecular results, all three methods produce correct predictions. 
However, for slightly more complex molecular structures, the specially trained GTR-Mol-VLM yields the most accurate predictions. Gemini 2.5 Pro incorrectly predicts all double and single bonds in the benzene ring as aromatic bonds—though chemically correct, this is erroneous in graph-based evaluations. Finally, GPT-5 exhibits the greatest discrepancy in its predictions.

\subsection{Graph Qualitative Results}

\begin{figure*}
    \centering
    \includegraphics[width=1\linewidth]{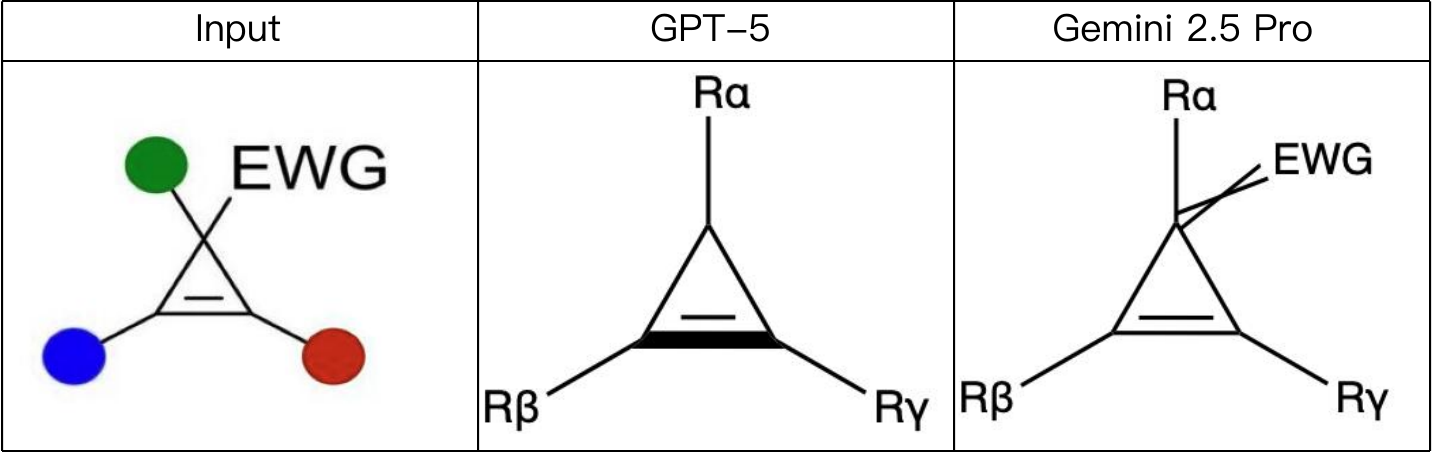}
    \caption{Comparison of Graph results produced by different models.}
    \label{fig:graph}
\end{figure*}

As shown in Fig.\ref{fig:graph}, we present the prediction results of GPT-5 and Gemini 2.5 Pro in Graph. 
Graph is the most challenging evaluation protocol in MolRecBench-Wild.
Although GPT-5 and Gemini 2.5 Pro demonstrate strong instruction adherence when converting circular shapes, they exhibit poor predictive capabilities for diverse chemical bond types.

\section{MOSAIC}
\label{apx:b}

The visual difficulty types are reported in Tables \ref{tab:table_visual_1_to_12} and \ref{tab:table_visual_13_to_18}, and the chemical difficulty types are reported in Tables \ref{tab:table_chemical_1_to_12} and \ref{tab:table_chemical_13_to_24}. 
The visual-difficulty examples demonstrate that our dataset encompasses a wide range of appearance-level perturbations, substantially increasing the challenge of accurately recognizing molecular diagrams. 
The chemical-difficulty examples further reveal that the dataset includes numerous uncommon or non-standard structural expression styles, which can markedly hinder a model’s ability to interpret molecules.
Collectively, these examples underscore that our benchmark provides substantially broader coverage of visual interference, symbolic variability, and long-tail structural edge cases than existing OCSR test sets, offering a more rigorous and fine-grained evaluation of model robustness in realistic chemical-document scenarios.

\newcommand{\hcimg}[1]{%
  \includegraphics[
    width=0.14\textwidth,
    height=0.14\textwidth,
    keepaspectratio
  ]{#1}%
}

\begin{table*}[t]
\caption{Example of visual dimension hard cases, ID 1--12. Each column group lists the hard case ID, type name, and a representative molecule image.}
\centering
\begingroup
\setlength{\tabcolsep}{6pt}
\renewcommand{\arraystretch}{2.9}
\begin{tabular}{|c|c|@{\hspace{20pt}}c|c|c|@{\hspace{20pt}}c|}
\hline
\textbf{ID} & \textbf{Type} & \textbf{Image} &
\textbf{ID} & \textbf{Type} & \textbf{Image} \\
\hline

\multirow{2}{*}{1} &
\multirow{2}{*}{{\color[HTML]{D83931} Decorated Text}} &
\multirow{2}{*}{\hcimg{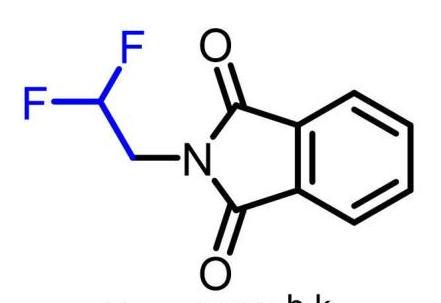}} &
\multirow{2}{*}{2} &
\multirow{2}{*}{{\color[HTML]{D83931} Decorated Bond}} &
\multirow{2}{*}{\hcimg{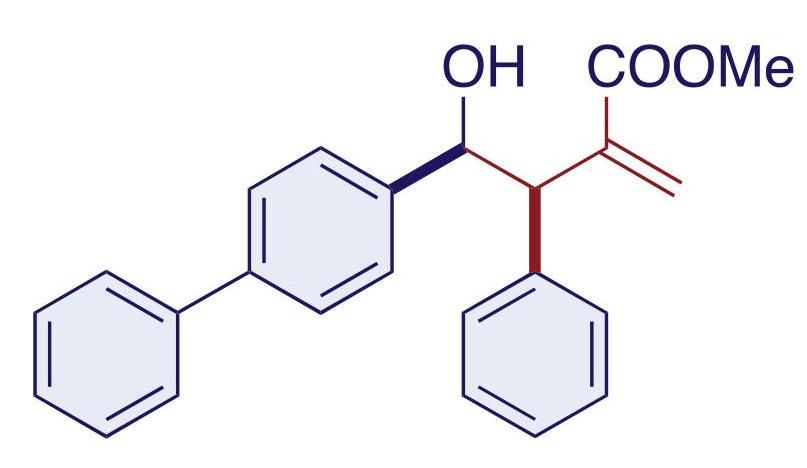}} \\
& & & & & \\ \hline

\multirow{2}{*}{3} &
\multirow{2}{*}{{\color[HTML]{D83931} Polluted Boundary}} &
\multirow{2}{*}{\hcimg{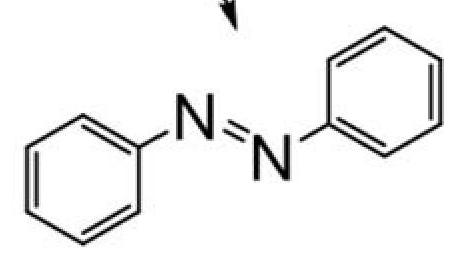}} &
\multirow{2}{*}{4} &
\multirow{2}{*}{{\color[HTML]{D83931} Blurry Image}} &
\multirow{2}{*}{\hcimg{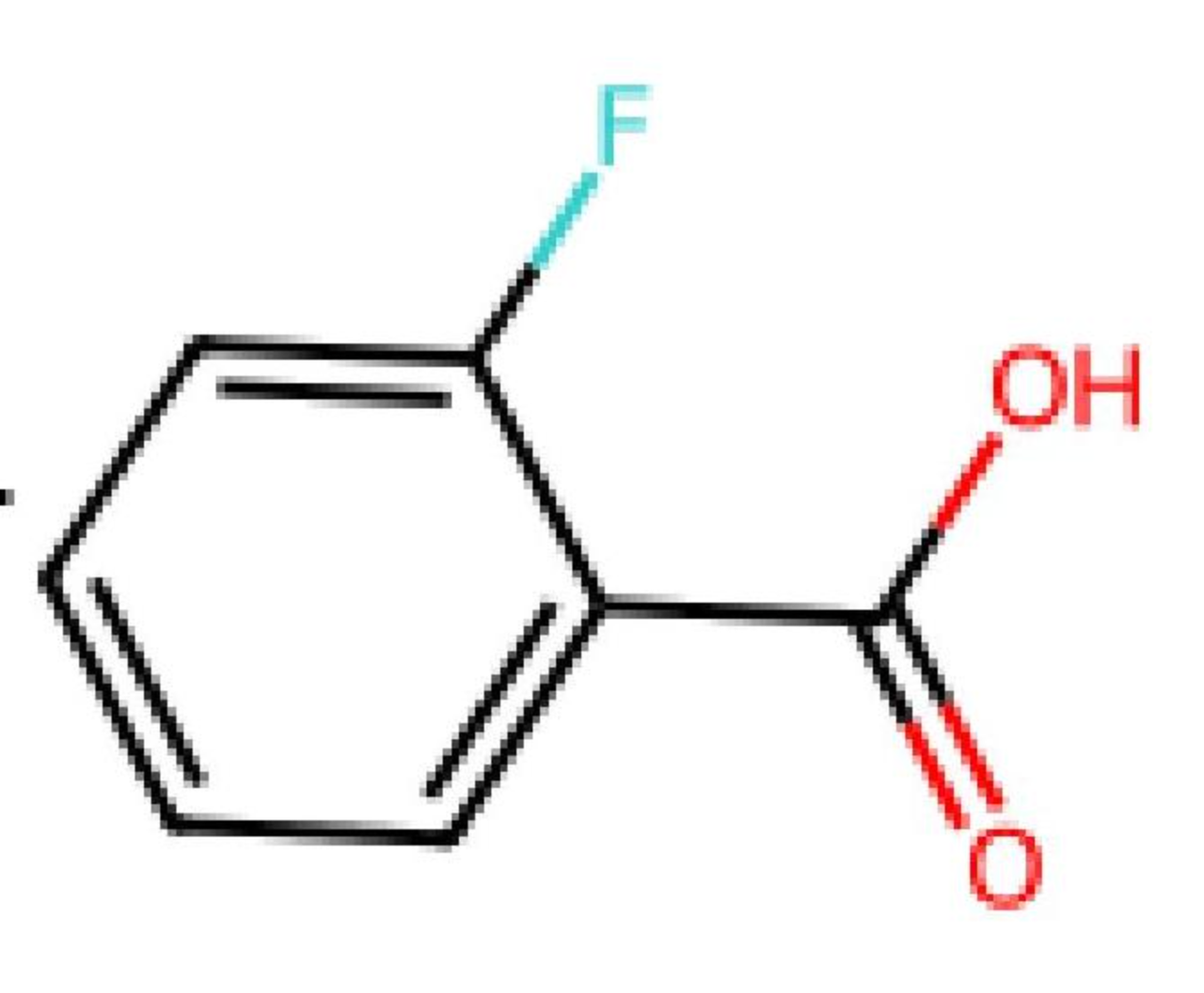}} \\
& & & & & \\ \hline

\multirow{2}{*}{5} &
\multirow{2}{*}{%
  \begin{tabular}{c}
    {\color[HTML]{D83931} Additional Arrow, }\\
    {\color[HTML]{D83931} Box, Text}
  \end{tabular}
} &
\multirow{2}{*}{\hcimg{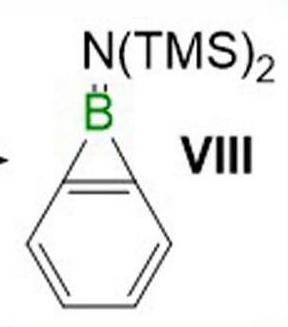}} &
\multirow{2}{*}{6} &
\multirow{2}{*}{%
  \begin{tabular}{c}
    {\color[HTML]{D83931} Colored Areas or}\\
    {\color[HTML]{D83931} Image Background}
  \end{tabular}
} &
\multirow{2}{*}{\hcimg{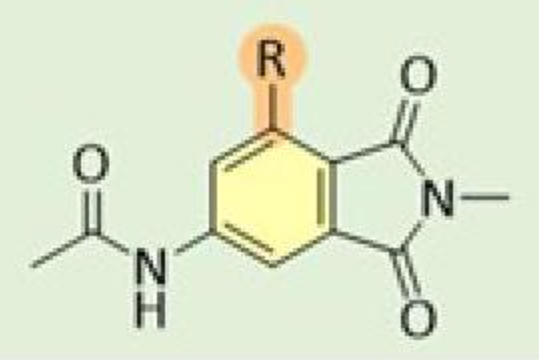}} \\
& & & & & \\ \hline

\multirow{2}{*}{7} &
\multirow{2}{*}{{\color[HTML]{D83931} Bond Crossing}} &
\multirow{2}{*}{\hcimg{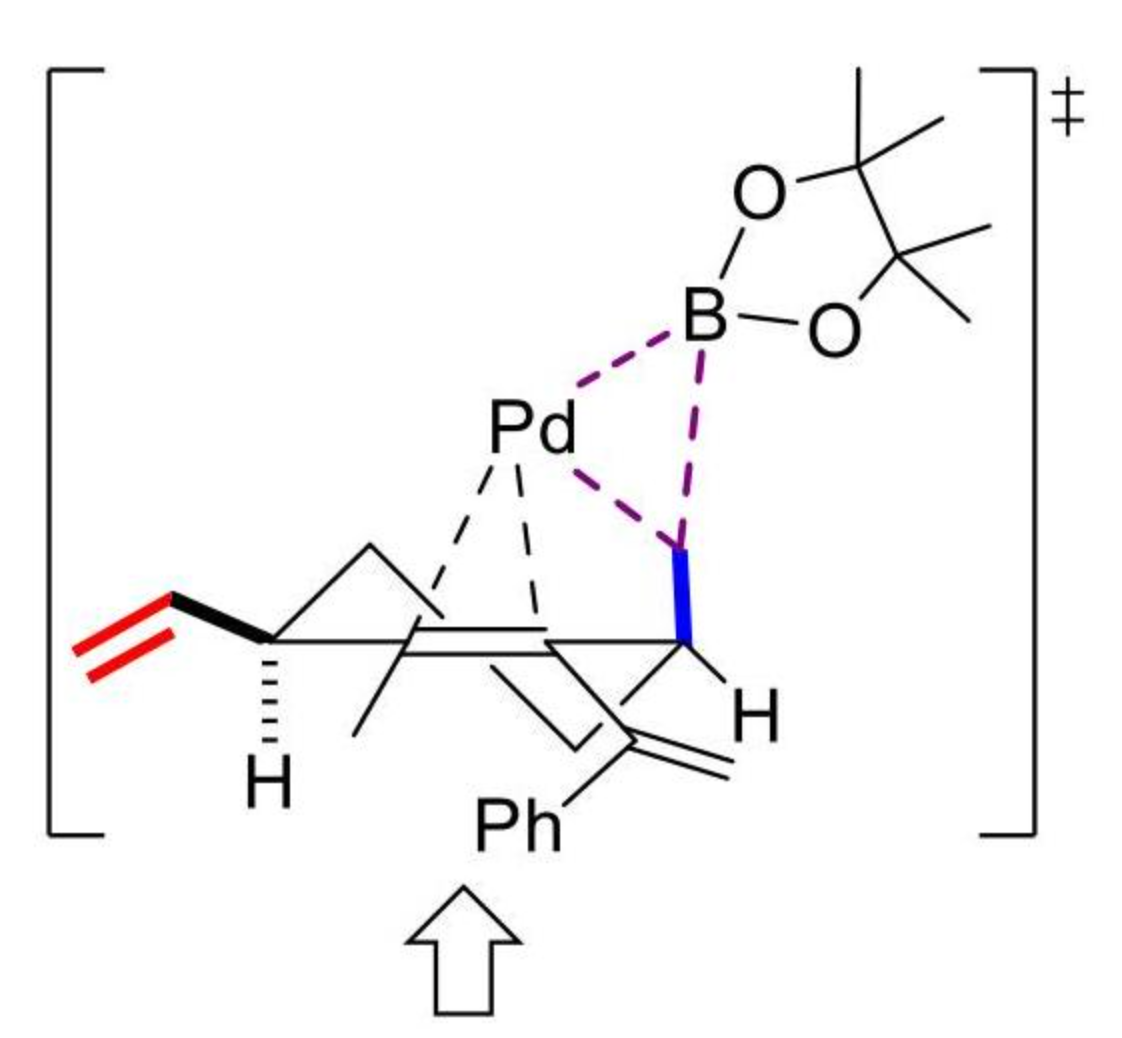}} &
\multirow{2}{*}{8} &
\multirow{2}{*}{%
  \begin{tabular}{c}
    {\color[HTML]{D83931} R Represented}\\
    {\color[HTML]{D83931} by Pattern}
  \end{tabular}
} &
\multirow{2}{*}{\hcimg{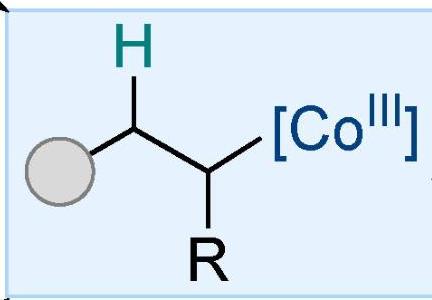}} \\
& & & & & \\ \hline

\multirow{2}{*}{9} &
\multirow{2}{*}{{\color[HTML]{D83931} Colored Ar}} &
\multirow{2}{*}{\hcimg{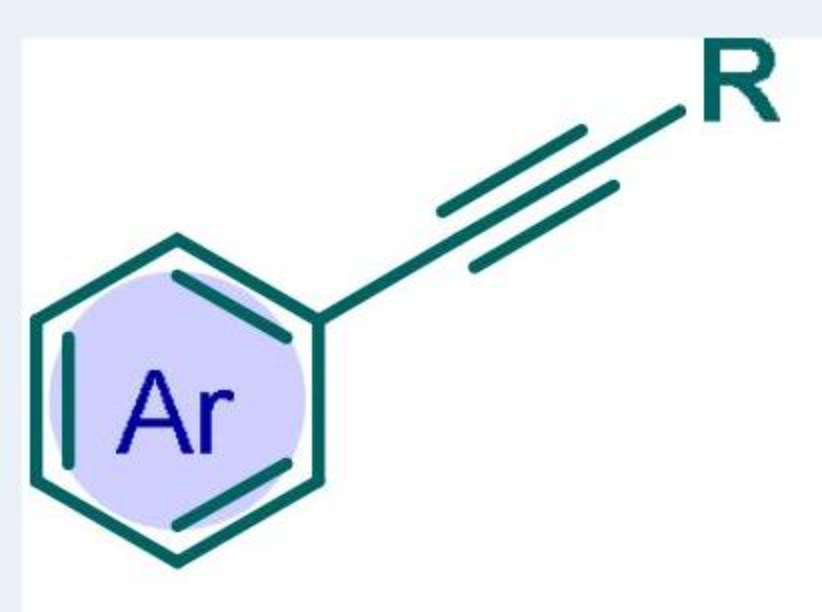}} &
\multirow{2}{*}{10} &
\multirow{2}{*}{{\color[HTML]{D83931} Short Bond}} &
\multirow{2}{*}{\hcimg{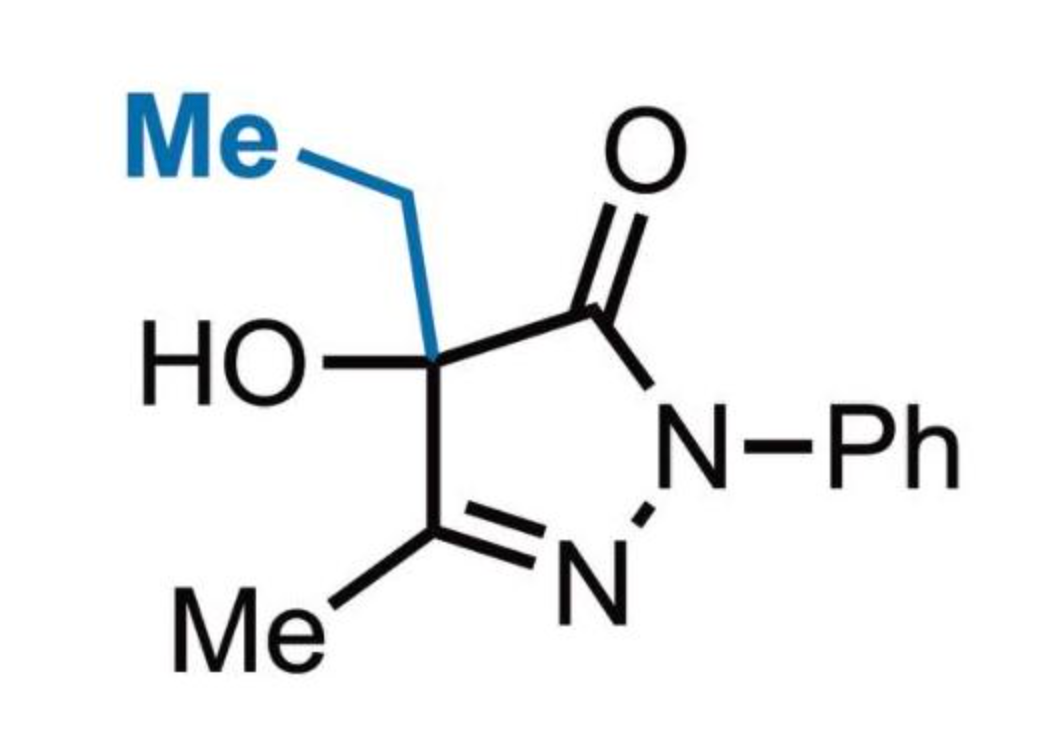}} \\
& & & & & \\ \hline

\multirow{2}{*}{11} &
\multirow{2}{*}{{\color[HTML]{D83931} Numbered Atom}} &
\multirow{2}{*}{\hcimg{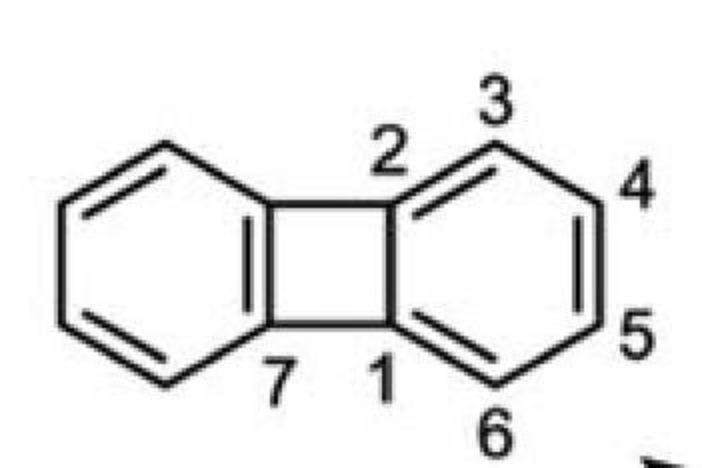}} &
\multirow{2}{*}{12} &
\multirow{2}{*}{{\color[HTML]{D83931} Incomplete Molecule}} &
\multirow{2}{*}{\hcimg{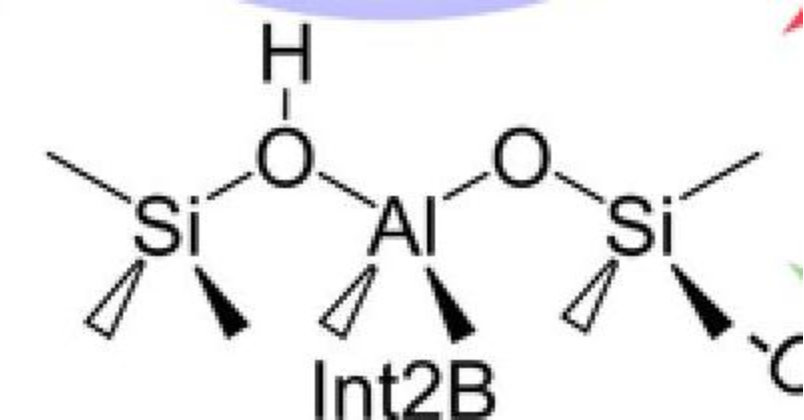}} \\
& & & & & \\ \hline

\end{tabular}
\endgroup
\label{tab:table_visual_1_to_12}
\end{table*}

\begin{table*}[t]
\caption{Example of visual dimension hard cases, ID 13--18.}
\centering
\begingroup
\setlength{\tabcolsep}{6pt}
\renewcommand{\arraystretch}{2.8}
\begin{tabular}{|c|c|@{\hspace{18pt}}c|c|c|@{\hspace{18pt}}c|}
\hline
\textbf{ID} & \textbf{Type} & \textbf{Image} &
\textbf{ID} & \textbf{Type} & \textbf{Image} \\
\hline

\multirow{2}{*}{13} &
\multirow{2}{*}{{\color[HTML]{D83931} Large Molecule}} &
\multirow{2}{*}{\hcimg{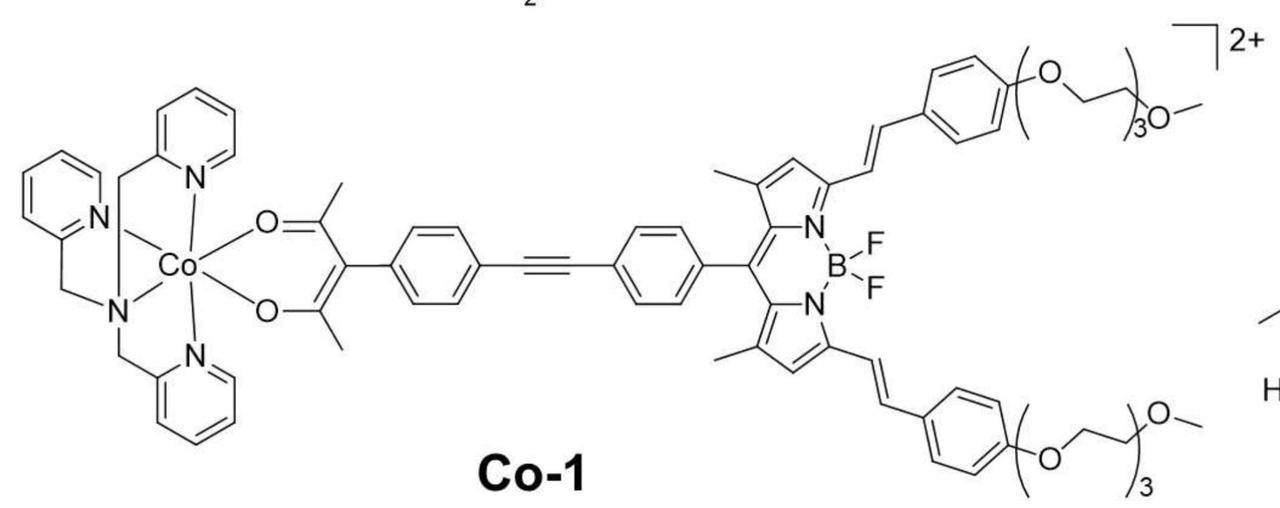}} &
\multirow{2}{*}{14} &
\multirow{2}{*}{{\color[HTML]{D83931} Large Font}} &
\multirow{2}{*}{\hcimg{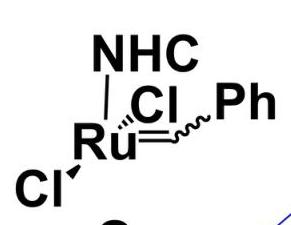}} \\
& & & & & \\ \hline

\multirow{2}{*}{15} &
\multirow{2}{*}{{\color[HTML]{D83931} Long Bond}} &
\multirow{2}{*}{\hcimg{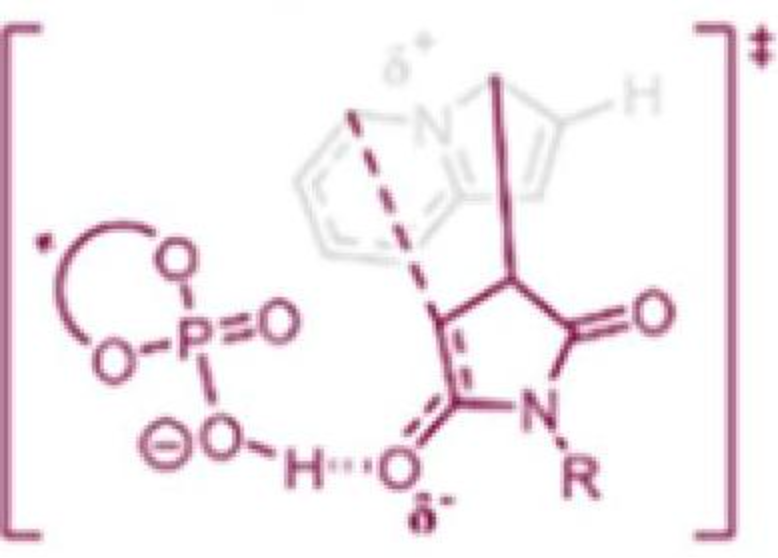}} &
\multirow{2}{*}{16} &
\multirow{2}{*}{{\color[HTML]{D83931} Thick Bond}} &
\multirow{2}{*}{\hcimg{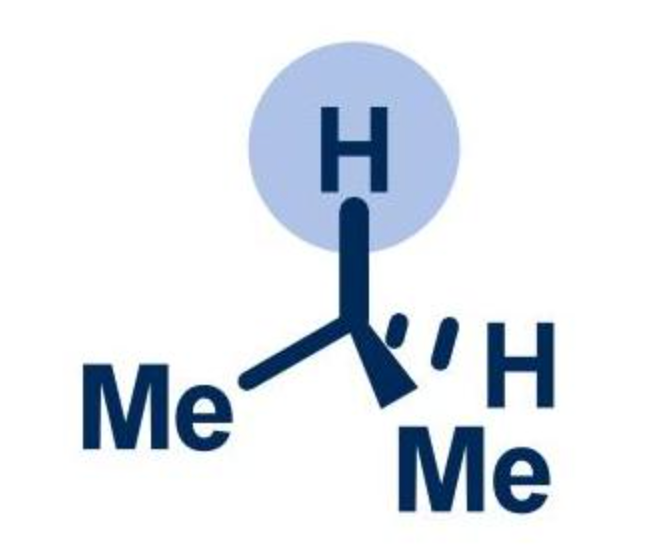}} \\
& & & & & \\ \hline

\multirow{2}{*}{17} &
\multirow{2}{*}{{\color[HTML]{D83931} Thin Bond}} &
\multirow{2}{*}{\hcimg{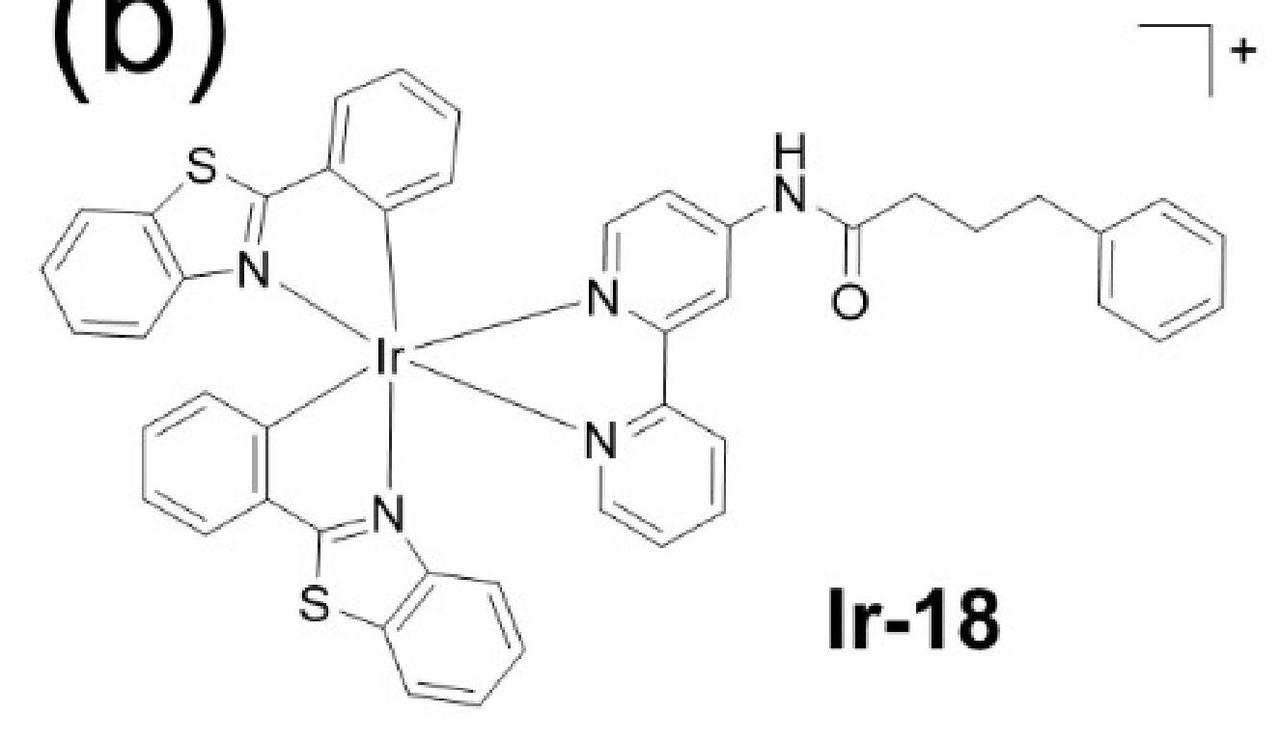}} &
\multirow{2}{*}{18} &
\multirow{2}{*}{%
  \begin{tabular}{c}
    {\color[HTML]{D83931} Long Functional}\\
    {\color[HTML]{D83931} Group Name}
  \end{tabular}
} &
\multirow{2}{*}{\hcimg{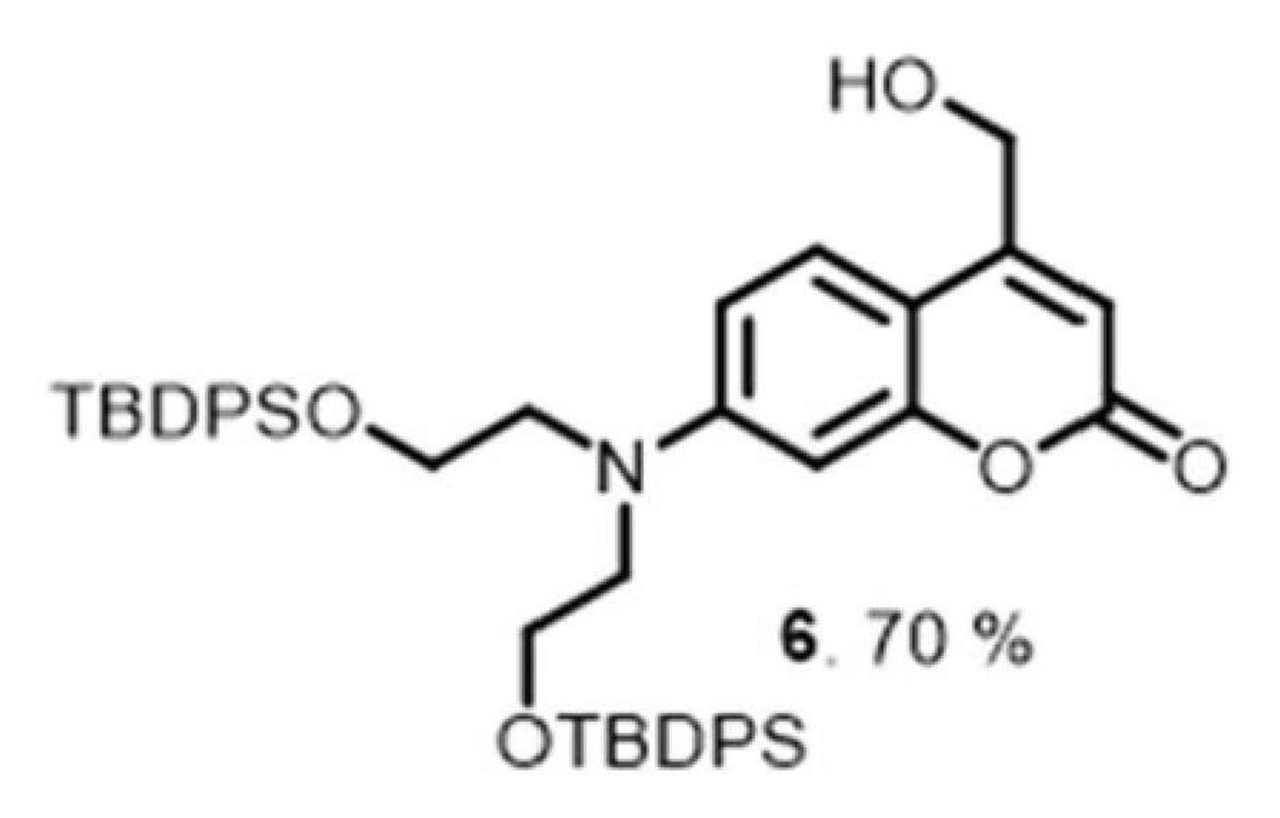}} \\
& & & & & \\ \hline

\end{tabular}
\endgroup
\label{tab:table_visual_13_to_18}
\end{table*}

\begin{table*}[t]
\caption{Example of chemical dimension hard cases, ID 1--12.}
\centering
\begingroup
\setlength{\tabcolsep}{6pt}
\renewcommand{\arraystretch}{2.6}
\begin{tabular}{|c|c|@{\hspace{18pt}}c|c|c|@{\hspace{8pt}}c|}
\hline
\textbf{ID} & \textbf{Type} & \textbf{Image} &
\textbf{ID} & \textbf{Type} & \textbf{Image} \\
\hline

\multirow{2}{*}{1} &
\multirow{2}{*}{{\color[HTML]{D83931} Equal-width Chiral Bond}} &
\multirow{2}{*}{\hcimg{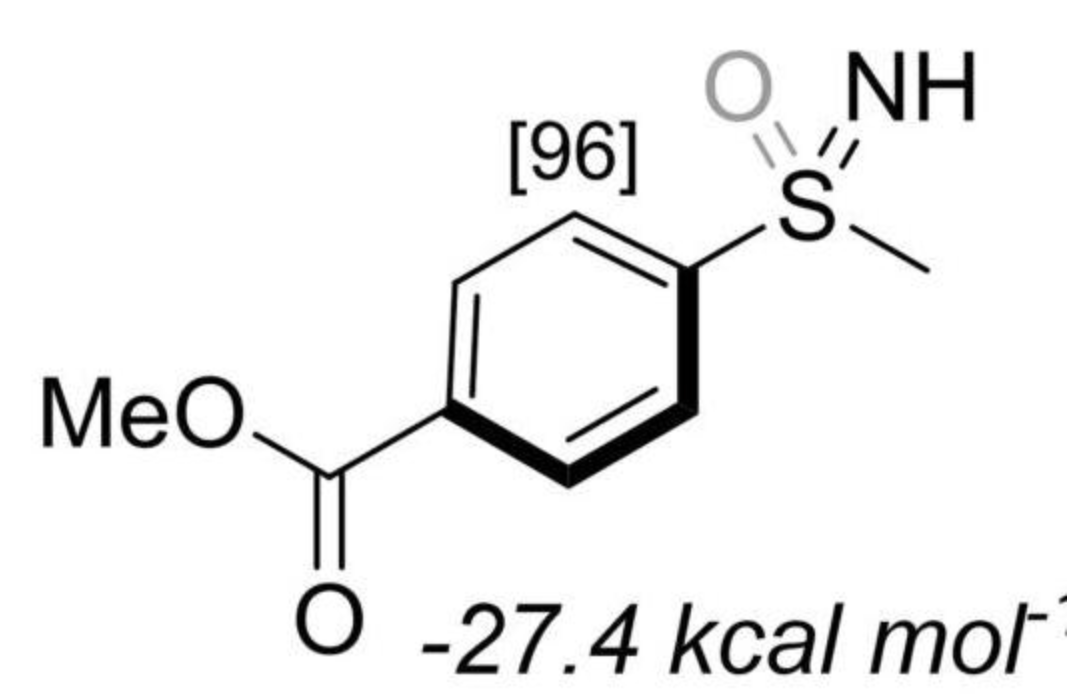}} &
\multirow{2}{*}{2} &
\multirow{2}{*}{{\color[HTML]{000000} Charge Symbol}} &
\multirow{2}{*}{\hcimg{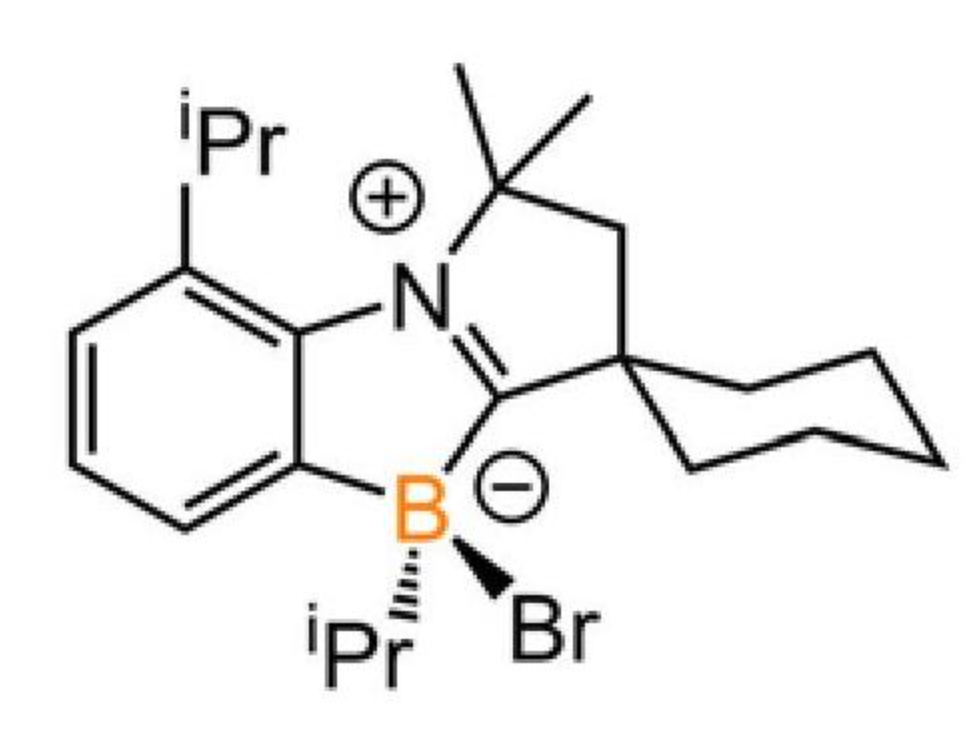}} \\
& & & & & \\ \hline

\multirow{2}{*}{3} &
\multirow{2}{*}{{\color[HTML]{000000} Dashed Bond}} &
\multirow{2}{*}{\hcimg{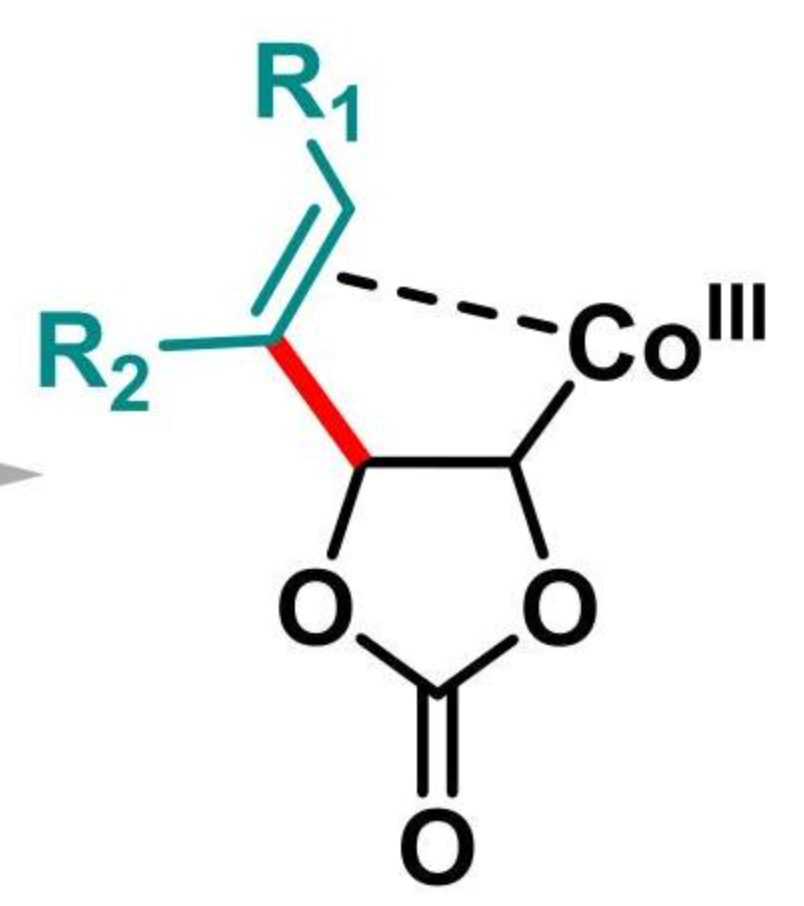}} &
\multirow{2}{*}{4} &
\multirow{2}{*}{{\color[HTML]{000000} Wavy Bond}} &
\multirow{2}{*}{\hcimg{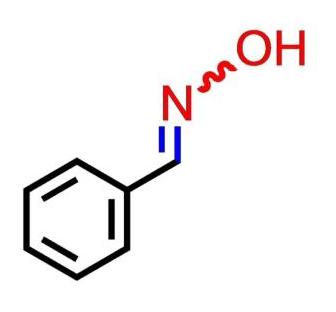}} \\
& & & & & \\ \hline

\multirow{2}{*}{5} &
\multirow{2}{*}{%
  \begin{tabular}{c}
    {\color[HTML]{D83931} Lone Pair}\\
    {\color[HTML]{D83931} Electron Symbol}
  \end{tabular}
} &
\multirow{2}{*}{\hcimg{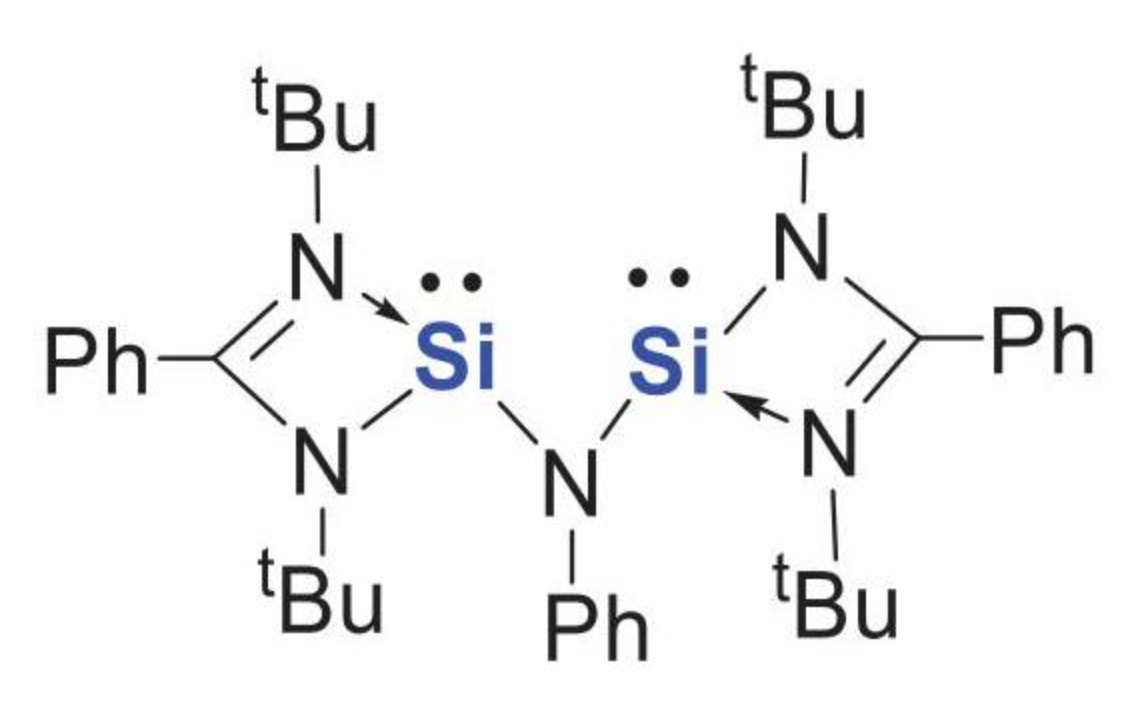}} &
\multirow{2}{*}{6} &
\multirow{2}{*}{{\color[HTML]{000000} Triple Bond}} &
\multirow{2}{*}{\hcimg{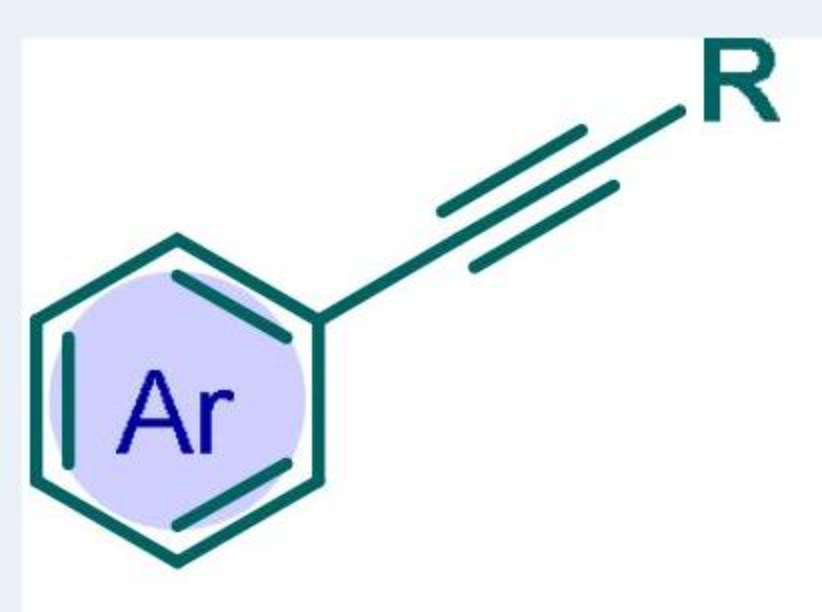}} \\
& & & & & \\ \hline

\multirow{2}{*}{7} &
\multirow{2}{*}{{\color[HTML]{000000} Hash Bond}} &
\multirow{2}{*}{\hcimg{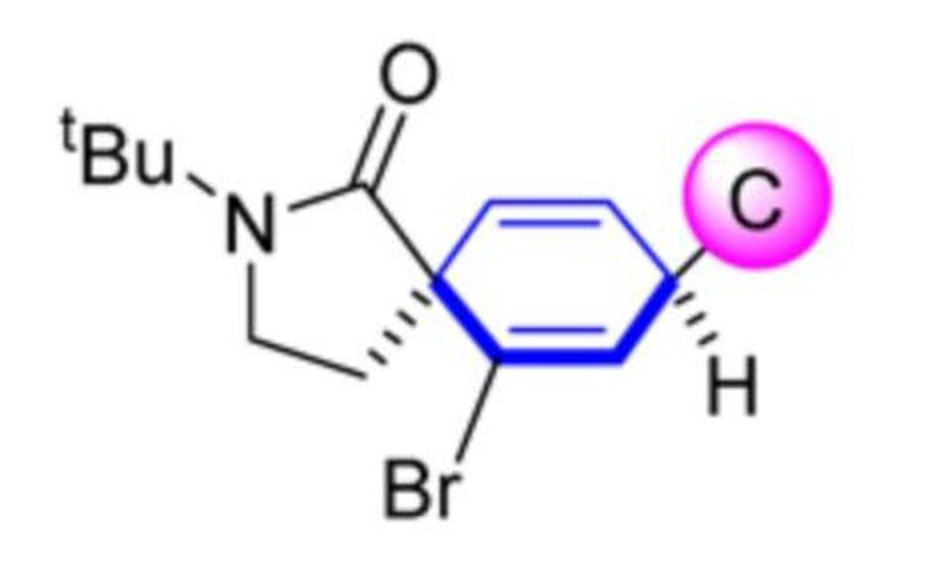}} &
\multirow{2}{*}{8} &
\multirow{2}{*}{{\color[HTML]{000000} Ionic Bond}} &
\multirow{2}{*}{\hcimg{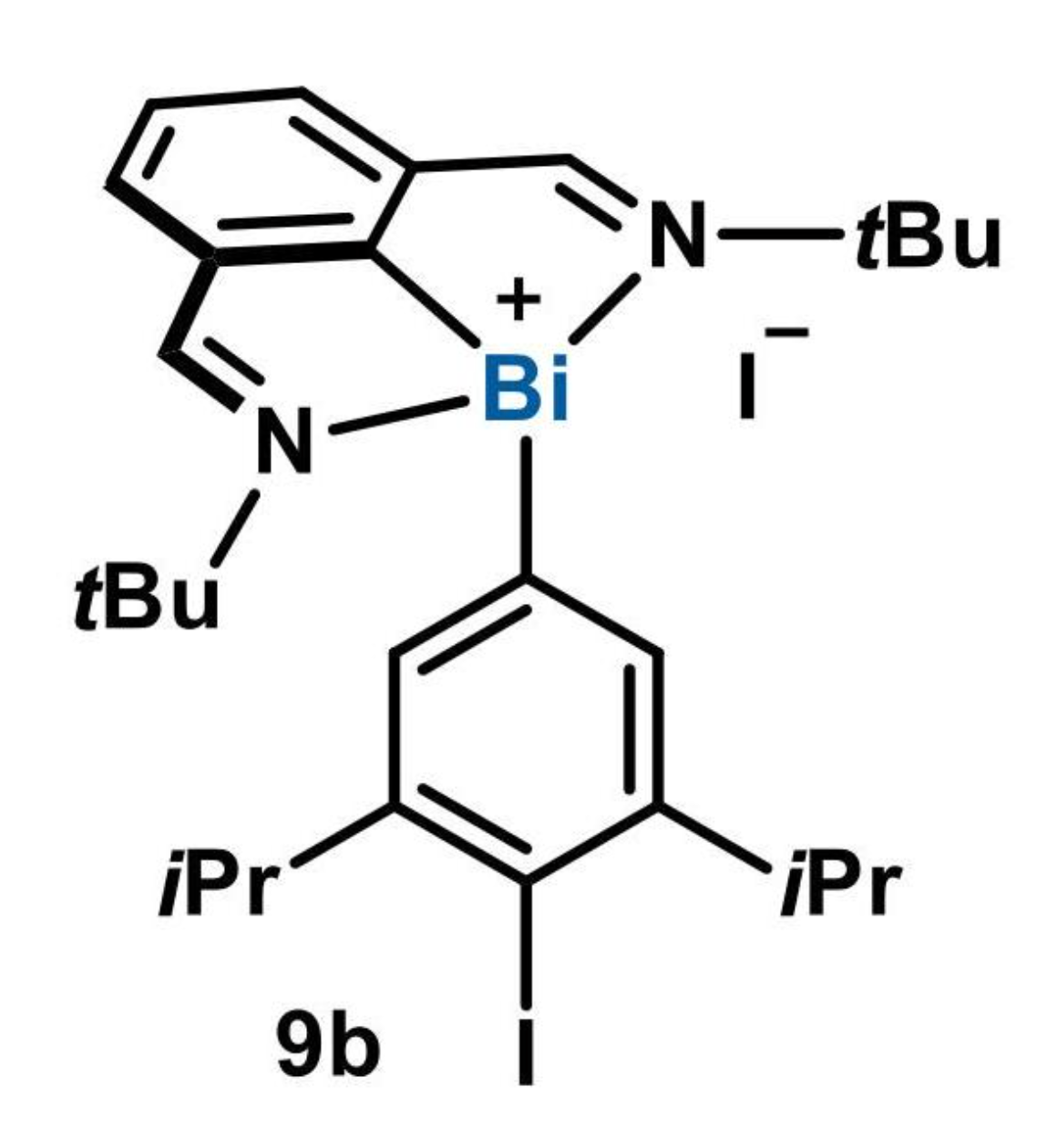}} \\
& & & & & \\ \hline

\multirow{2}{*}{9} &
\multirow{2}{*}{%
  \begin{tabular}{c}
    {\color[HTML]{D83931} R on Ar with}\\
    {\color[HTML]{D83931} uncertain position}
  \end{tabular}
} &
\multirow{2}{*}{\hcimg{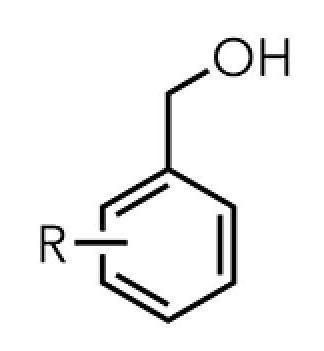}} &
\multirow{2}{*}{10} &
\multirow{2}{*}{{\color[HTML]{D83931} Abbreviated structure}} &
\multirow{2}{*}{\hcimg{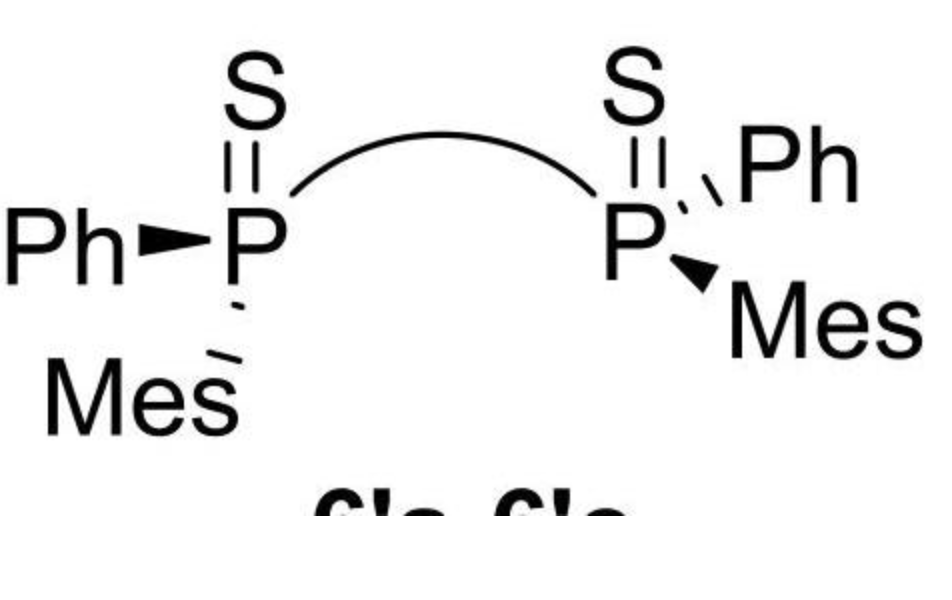}} \\
& & & & & \\ \hline

\multirow{2}{*}{11} &
\multirow{2}{*}{{\color[HTML]{000000} Valence Symbol}} &
\multirow{2}{*}{\hcimg{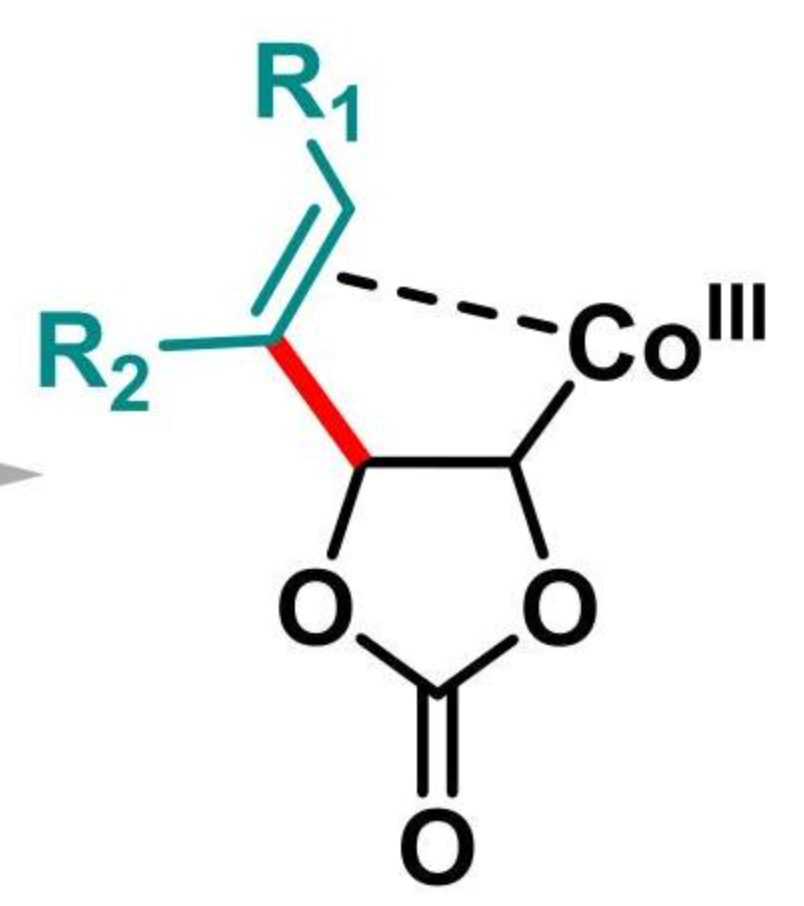}} &
\multirow{2}{*}{12} &
\multirow{2}{*}{{\color[HTML]{000000} Polymer}} &
\multirow{2}{*}{\hcimg{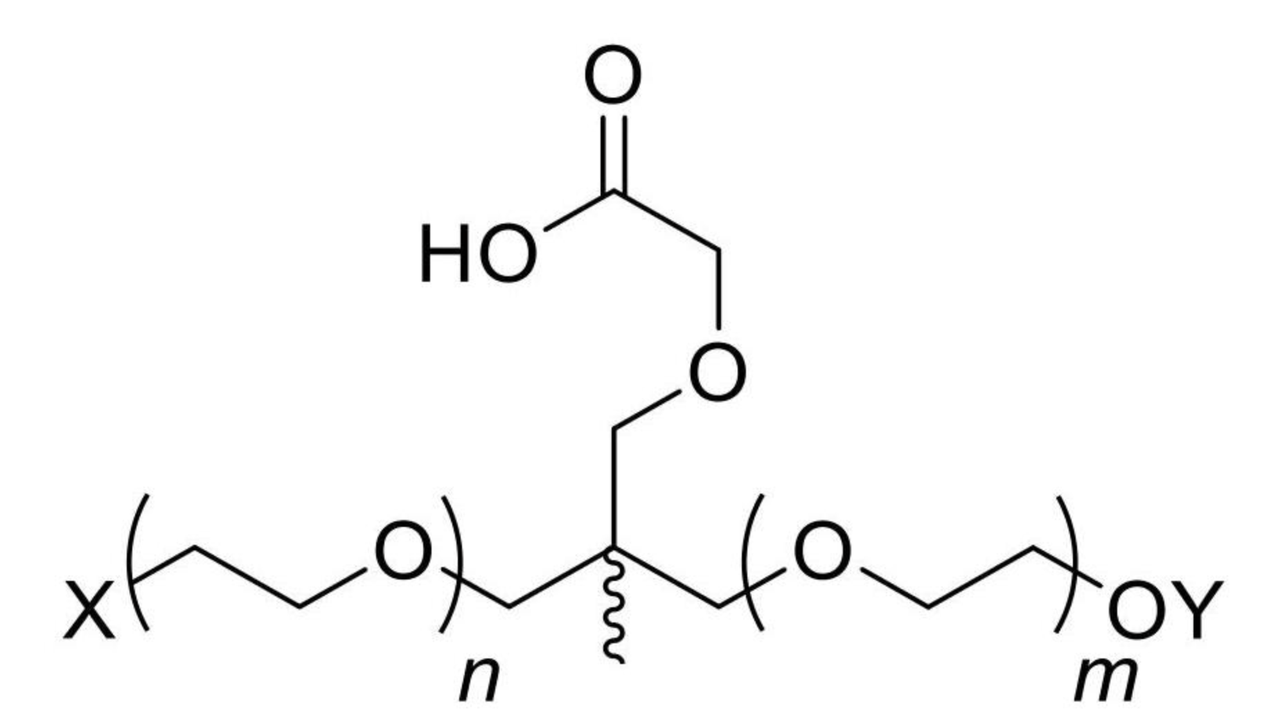}} \\
& & & & & \\ \hline

\end{tabular}
\endgroup
\label{tab:table_chemical_1_to_12}
\end{table*}

\begin{table*}[t]
\caption{Example of chemical dimension hard cases, ID 13--24.}
\centering
\begingroup
\setlength{\tabcolsep}{6pt}
\renewcommand{\arraystretch}{2.6}
\begin{tabular}{|c|c|@{\hspace{18pt}}c|c|c|@{\hspace{8pt}}c|}
\hline
\textbf{ID} & \textbf{Type} & \textbf{Image} &
\textbf{ID} & \textbf{Type} & \textbf{Image} \\
\hline

\multirow{2}{*}{13} &
\multirow{2}{*}{{\color[HTML]{D83931} Aromatic Bond}} &
\multirow{2}{*}{\hcimg{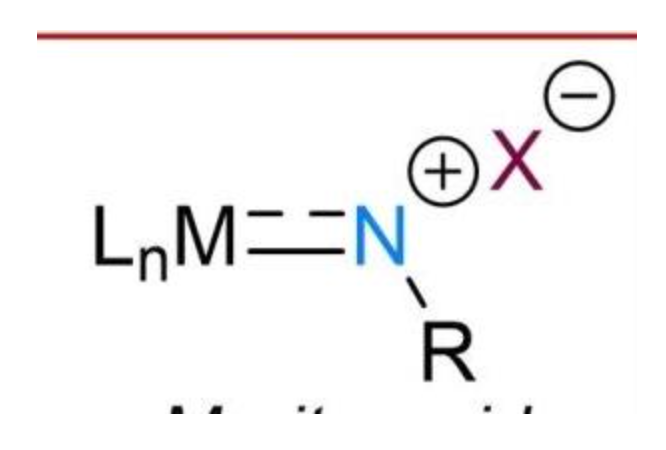}} &
\multirow{2}{*}{14} &
\multirow{2}{*}{{\color[HTML]{000000} Multi-group}} &
\multirow{2}{*}{\hcimg{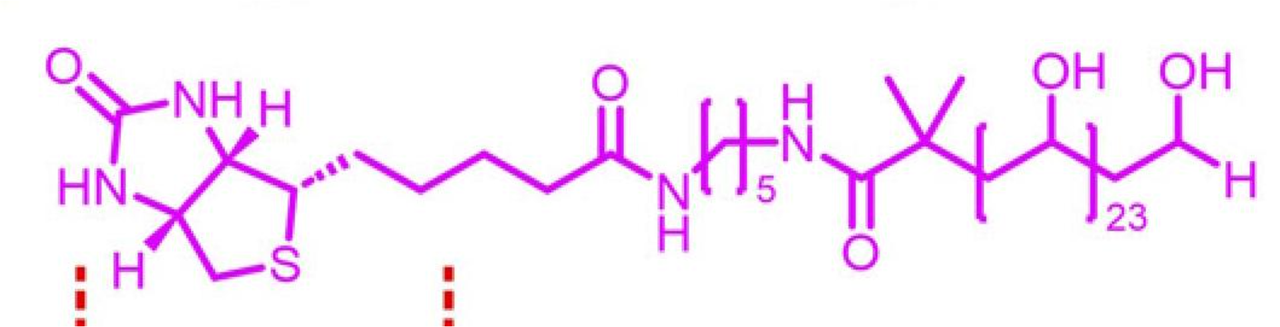}} \\
& & & & & \\ \hline

\multirow{2}{*}{15} &
\multirow{2}{*}{%
  \begin{tabular}{c}
    {\color[HTML]{D83931} Atom on Ar with}\\
    {\color[HTML]{D83931} uncertain position}
  \end{tabular}
} &
\multirow{2}{*}{\hcimg{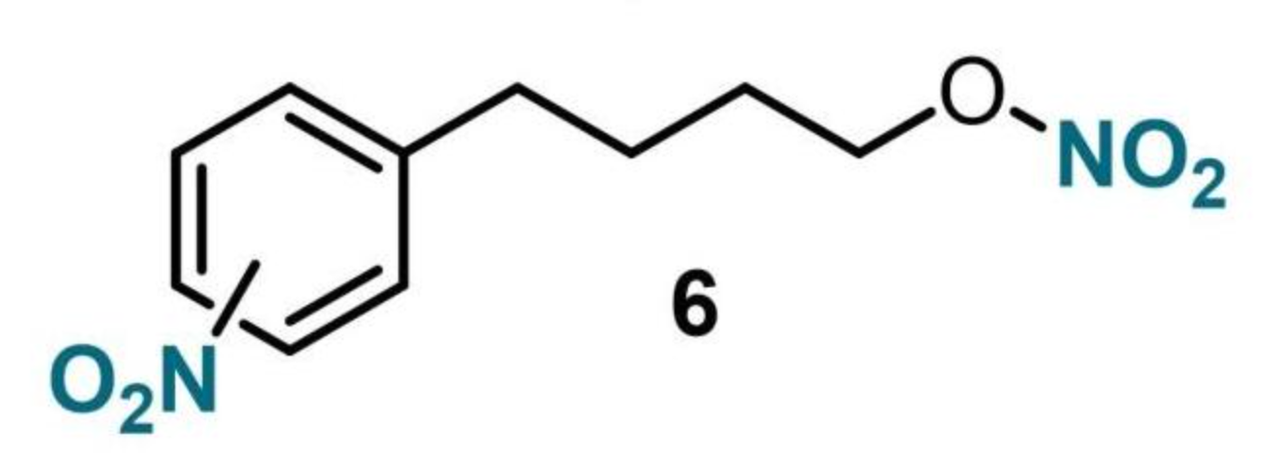}} &
\multirow{2}{*}{16} &
\multirow{2}{*}{{\color[HTML]{000000} Transition State}} &
\multirow{2}{*}{\hcimg{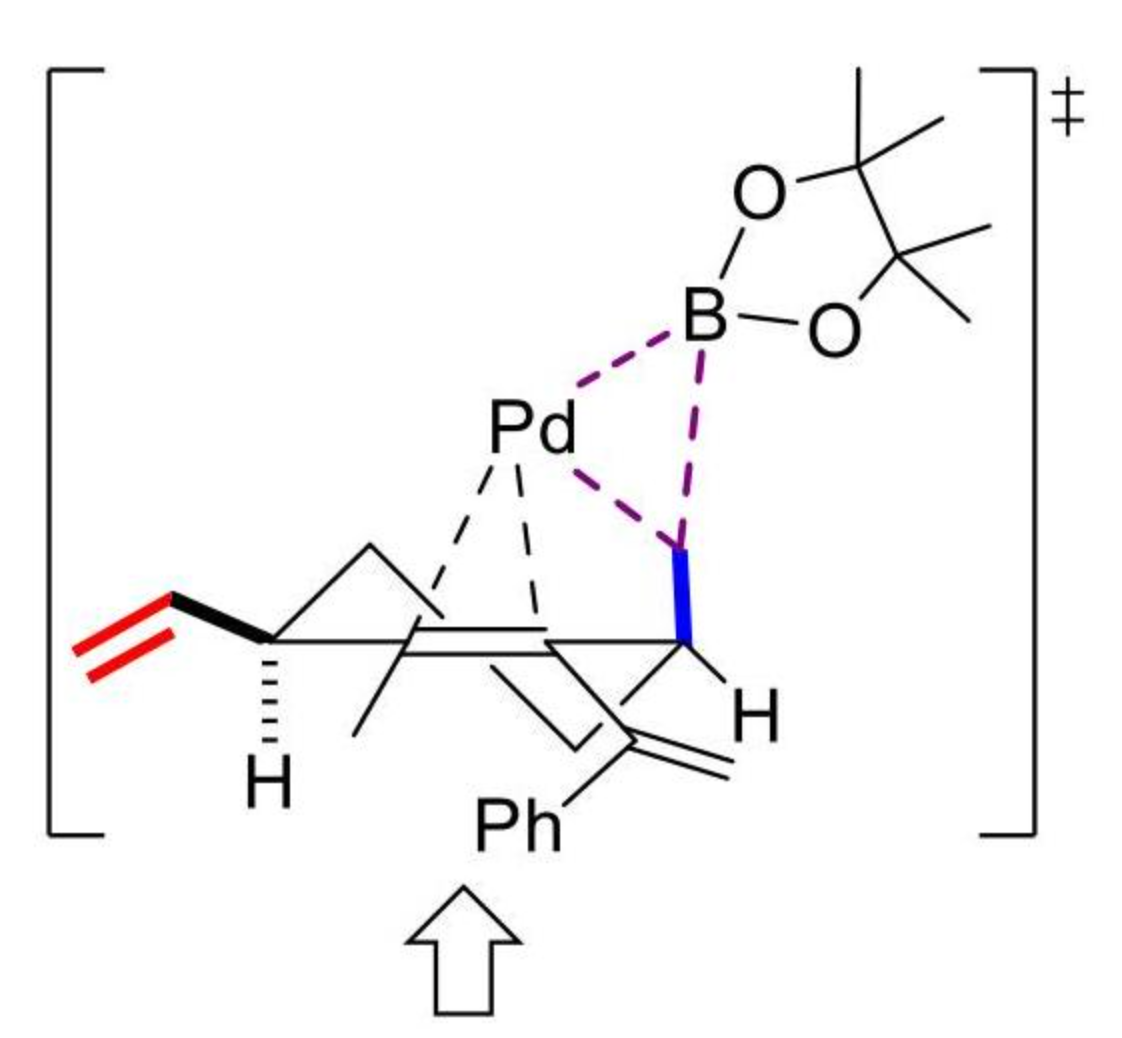}} \\
& & & & & \\ \hline

\multirow{2}{*}{17} &
\multirow{2}{*}{{\color[HTML]{000000} Coordination Bond}} &
\multirow{2}{*}{\hcimg{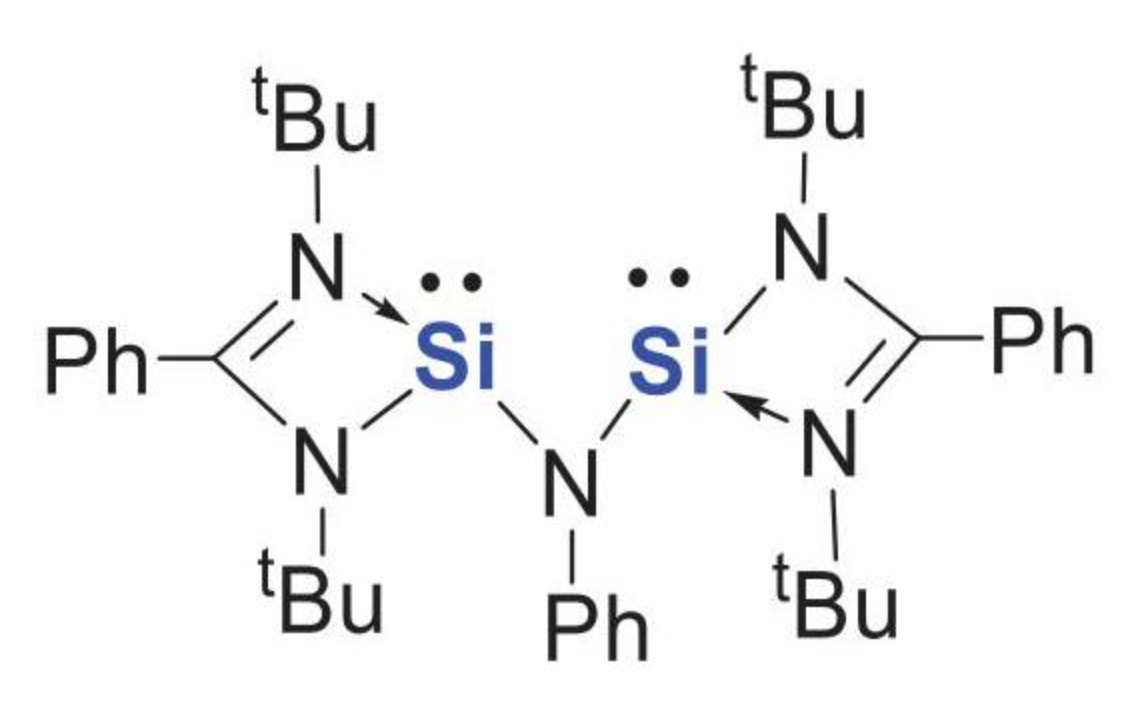}} &
\multirow{2}{*}{18} &
\multirow{2}{*}{%
  \begin{tabular}{c}
    {\color[HTML]{D83931} Consecutive}\\
    {\color[HTML]{D83931} Double Bond}
  \end{tabular}
} &
\multirow{2}{*}{\hcimg{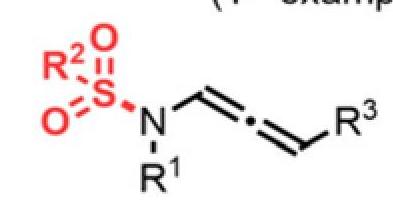}} \\
& & & & & \\ \hline

\multirow{2}{*}{19} &
\multirow{2}{*}{{\color[HTML]{000000} Double Dashed Bond}} &
\multirow{2}{*}{\hcimg{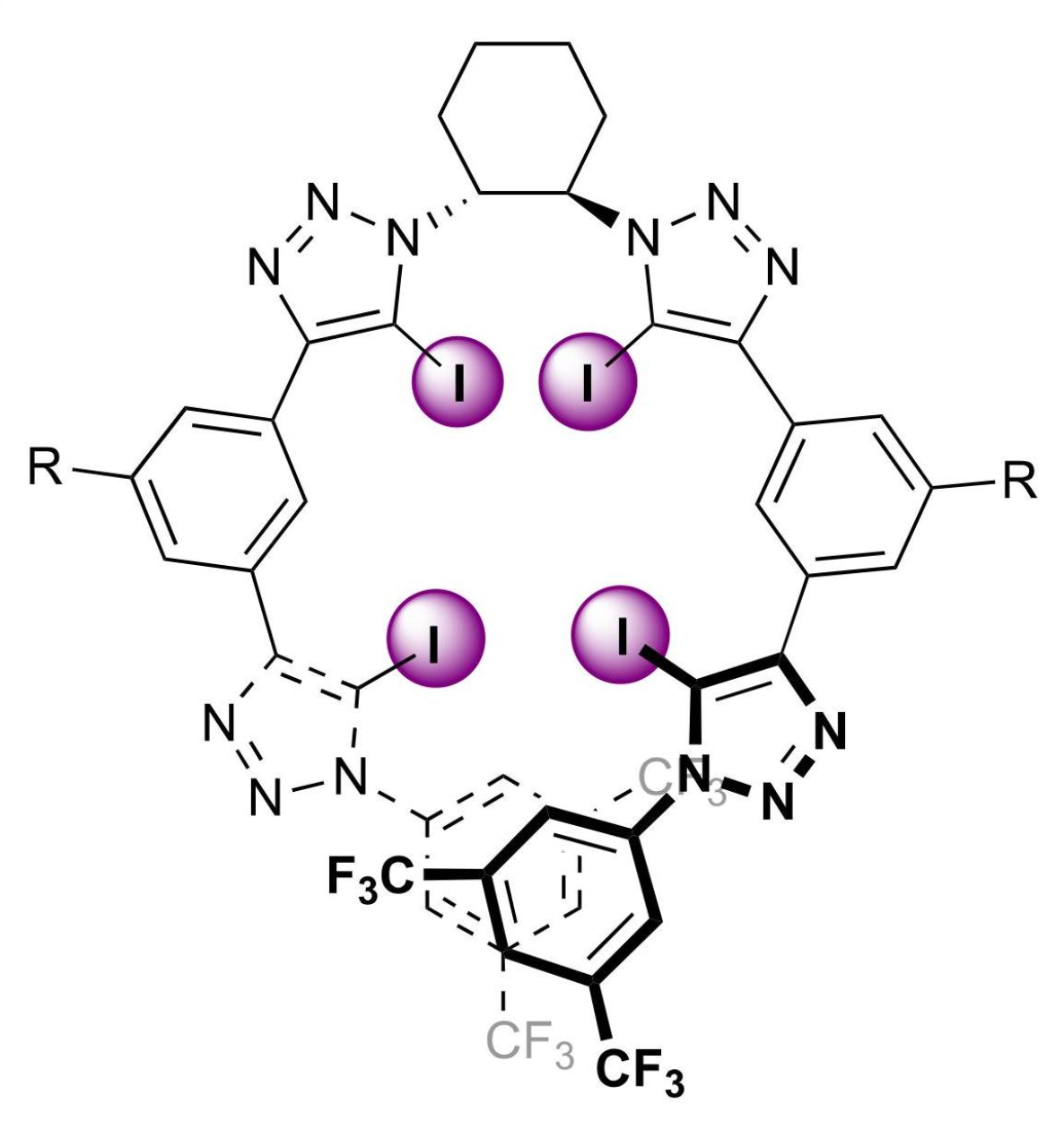}} &
& & \\
& & & & & \\ \hline

\end{tabular}
\endgroup
\label{tab:table_chemical_13_to_24}
\end{table*}

\section{Statistics}
\label{apx:c}
  
\begin{figure*}
    \centering
    \includegraphics[width=1.0\linewidth]{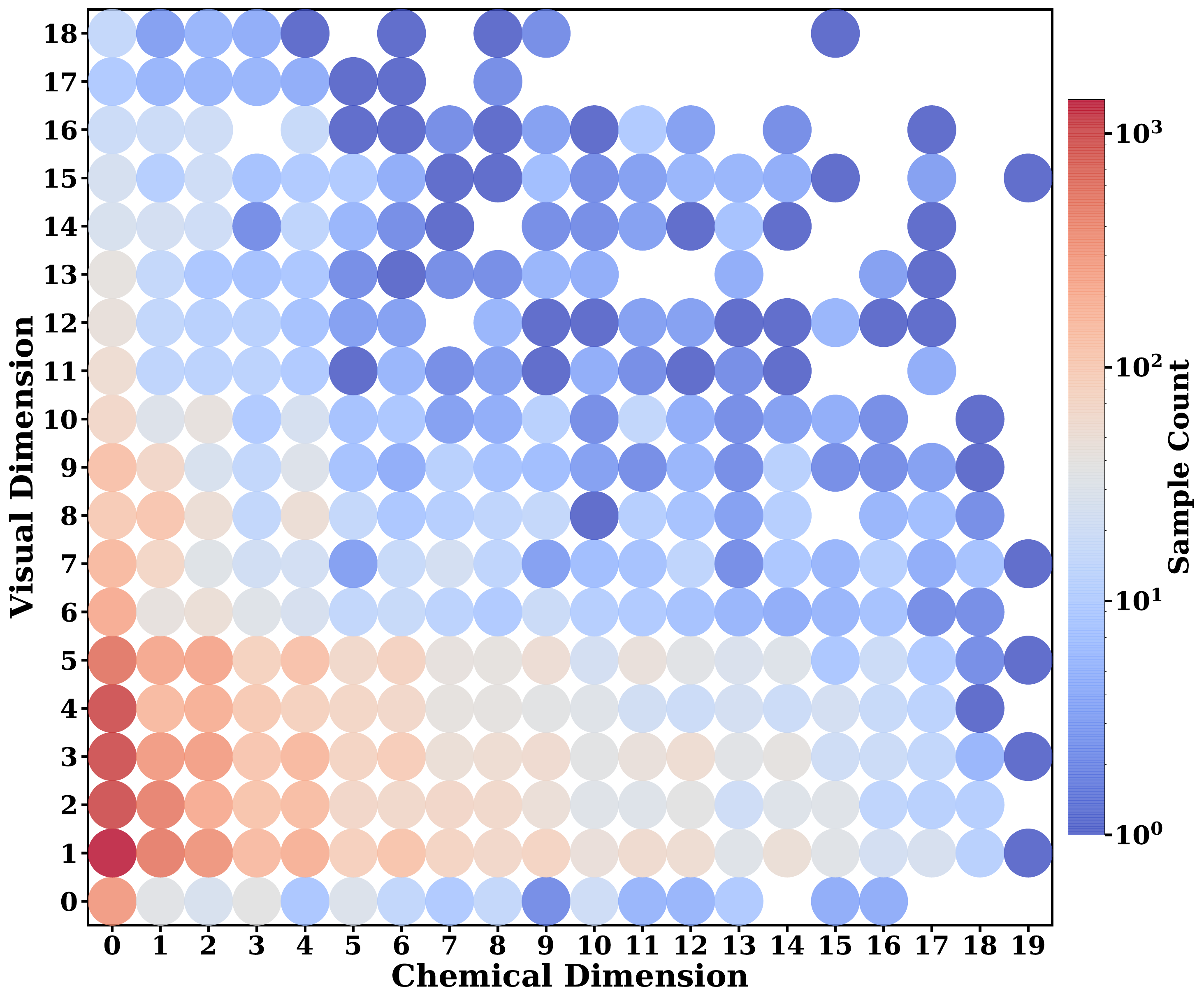}
    \caption{Joint heatmap showing the co-occurrence counts between each visual difficulty dimension and each chemical difficulty dimension in our benchmark. The horizontal axis corresponds to the indices of the chemical difficulty labels, and the vertical axis corresponds to the indices of the visual difficulty labels. The color intensity encodes the number of samples for each pair: warmer tones (reds) indicate higher counts, while cooler tones (blues) indicate lower counts.}
    \label{fig:heatmap_number_dimension}
\end{figure*}

For each sample, we annotated the types of visual and chemical difficulty labels and counted the number of combinations of different difficulty labels that appeared. 
As shown in Fig.~\ref{fig:heatmap_number_dimension}, each circle represents the number of samples that exhibit both types of difficulties simultaneously.
This distribution offers a more fine-grained assessment of model robustness under varying levels of visual and chemical complexity.
  
\section{Prompt}
\label{apx:d}

\begin{figure*}
    \centering
    \includegraphics[width=1.0\linewidth]{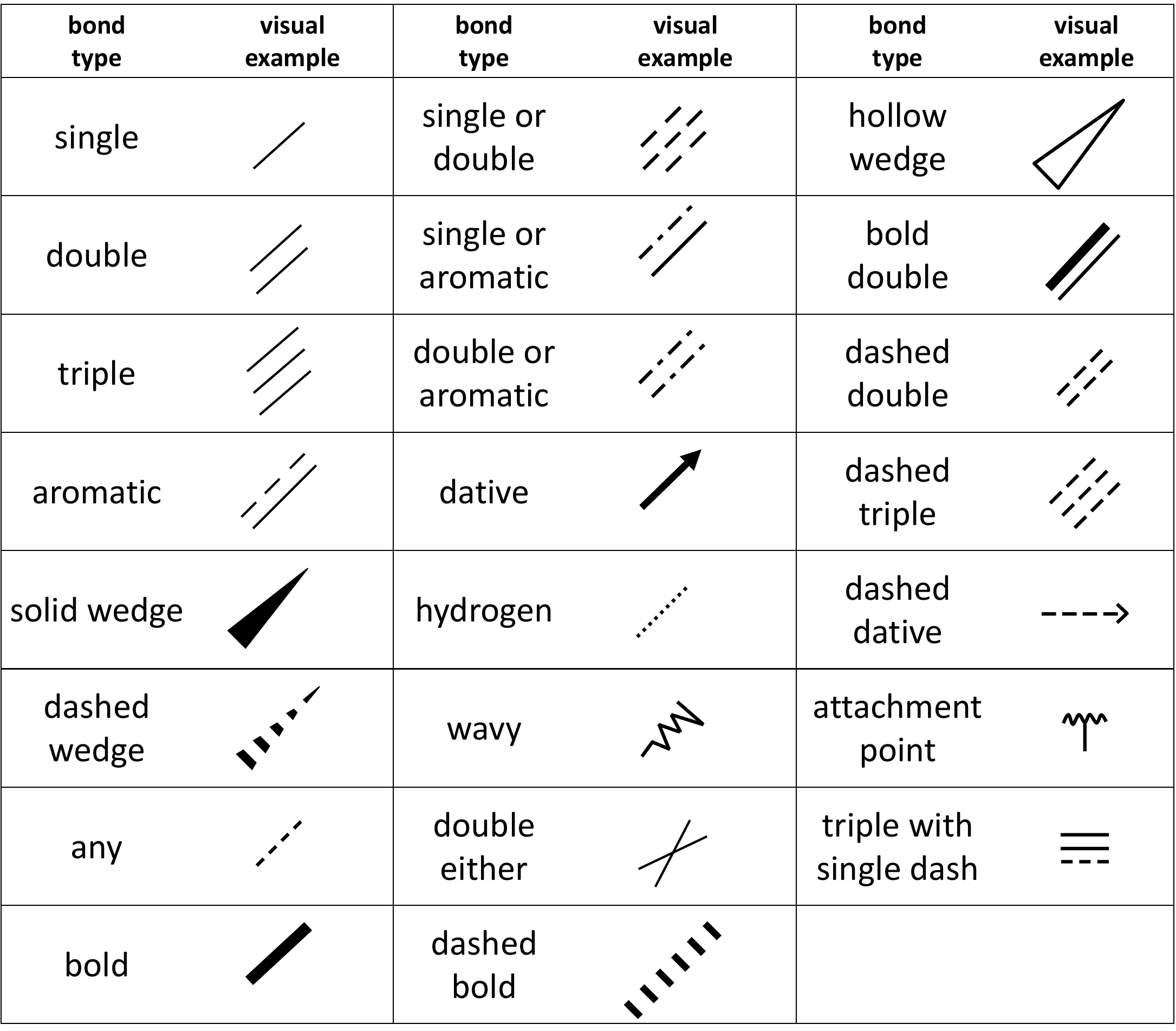}
    \caption{Visual appearance of different bonds.}
    \label{fig:bond_examplar}
\end{figure*}

\begin{figure*}
    \centering
    \includegraphics[width=1.0\linewidth]{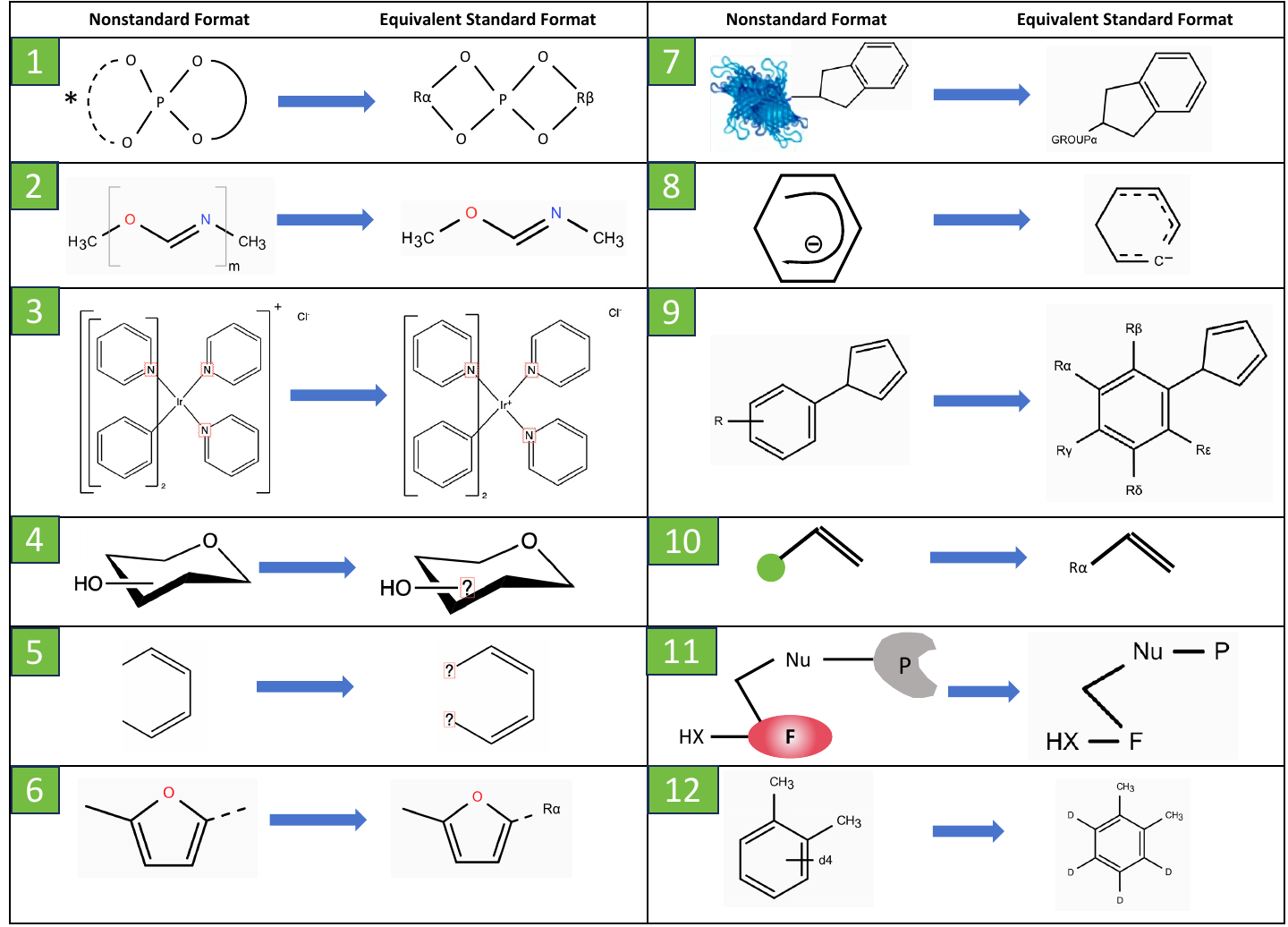}
    \caption{Different cases and their corresponding standardized forms.}
    \label{fig:case_examplar}
\end{figure*}

We designed specific prompt templates for each prediction task, as detailed below:
\begin{itemize}
\item \textbf{SMILES Prediction:} The corresponding prompt template is shown in Figure~\ref{fig:prompt_for_smiles}.
\item \textbf{Simplified Graph Prediction:} The corresponding prompt template is shown in Figure~\ref{fig:prompt_for_simplified_graph}.
\item \textbf{Full Graph Prediction:} The prompt template for this task, shown in Figure~\ref{fig:prompt_for_graph}, combines two visual prompts (Figure~\ref{fig:bond_examplar} and Figure~\ref{fig:case_examplar}).
\end{itemize}

\begin{figure*}[!ht]
\begin{AIbox}{Prompt template for the SMILES}
    {
 You are an expert chemist viewing a diagram of a chemical molecular structure. Your task is to generate the corresponding SMILES string for the molecule(s) shown in the image.

Follow these rules strictly:
    {Follow these rules strictly:}
    \begin{itemize}[leftmargin=*, label=\textbullet, itemsep=2pt]
\item Output Format: Present the result in a single JSON format. The JSON object should contain one key, \texttt{smiles}.
\item Abbreviations: If there are chemical formula abbreviations or groups written together (e.g., MeO, Et, Ph, Boc), treat them as a single unit and enclose them in square brackets. For example, \texttt{MeO} becomes \texttt{[MeO]}.
\item  Stereochemistry: If stereochemistry is indicated with wedges or dashes, use @ or @@ to represent the chiral center. For double bonds with defined geometry, use / and \ to specify the E/Z (cis/trans) isomerism.
\item  Charges and Isotopes: Represent atomic charges (e.g., [O-], [NH4+]) and isotopes (e.g., [2H], [14C]) exactly as shown.
\item  Error Handling: If the image is unreadable or does not contain a chemical structure, the value for the ``smiles" key should be null and you must add an \texttt{error} key, like: \texttt{\{"smiles": null,"error": "Unable to recognize a chemical structure."\}}
    \end{itemize}
    
    {Here is an example of a successful conversion:}
    
\noindent 
\textbf{User:} \textit{\textless shot of an image showing 2-methoxypyridine-4-carbonitrile with stereochemistry\textgreater}
\vspace{1ex} 

\noindent
\textbf{Assistant:}
\begin{lstlisting}[style=json]
{
  "smiles": "[MeO]c1nc(C#N)cc(C)c1"
}
\end{lstlisting}
}
\end{AIbox}
\caption{Prompt template for the SMILES.}
\label{fig:prompt_for_smiles}
\end{figure*}

\begin{figure*}[!ht]
\begin{AIbox}{Prompt template for the simplified graph}
    {

You are an expert model specializing in the analysis of chemical diagrams. Your sole task is to view the provided image and convert the molecular structure into a graph representation, consisting of atoms and bonds.

Present the results as a single, complete JSON object.

Follow these instructions meticulously:
\begin{itemize}[leftmargin=*, label=\textbullet, itemsep=2pt]
\item JSON Structure: The final JSON object must contain exactly two top-level keys:
  \begin{itemize}
      \item \texttt{"atoms"}: A list of objects, where each object represents one atom.
      \item \texttt{"bonds"}: A list of objects, where each object represents one bond.
    \end{itemize}
    
\item Atom Attributes: Each object in the "atoms" list must contain three keys:
      \begin{itemize}
      \item \texttt{"id"}: A unique integer identifier for the atom, starting from 0.
      \item \texttt{"atom"}: The atom's symbol (e.g., C, O, N). If an abbreviation or group is shown (e.g., Ph, Boc), treat it as a single unit and represent it as a string (e.g., \texttt{[Ph]}).
      \item \texttt{"point\_2d"}: The \texttt{[x, y]} coordinates of the atom in the image.
    \end{itemize}

\item Bond Attributes: Each object in the "bonds" list must contain three keys:
  \begin{itemize}
  \item \texttt{"atom1"} and \texttt{"atom2"}: The integer ids of the two atoms connected by the bond.

  \item \texttt{"bond\_type"}: The type of the bond, which must be one of the following strings: \texttt{single}, \texttt{double}, \texttt{triple}, \texttt{aromatic}, \texttt{solid wedge}, or \texttt{dashed wedge}.  For solid wedge and dashed wedge bonds, the direction is from \texttt{atom1} to \texttt{atom2}. A solid wedge means \texttt{atom2} is pointing out of the plane, while a dashed wedge means \texttt{atom2} is pointing into the plane.
\end{itemize}

\item Strict Formatting: You must output only the single JSON object and nothing else. Do not include any additional text, explanations, or formatting like markdown code fences.
  \end{itemize}
    {Here is an example of a successful conversion:}
    
\noindent 
\textbf{User:} \textit{\textless shot of a chemical molecular formula image\textgreater}
\vspace{1ex} 

\noindent
\textbf{Assistant:}
  \begin{lstlisting}[style=json]
{
  "atoms": [
    {"id": 0, "atom": "C", "point_2d": [150, 200]},
    {"id": 1, "atom": "O", "point_2d": [250, 200]},
    {"id": 2, "atom": "N", "point_2d": [150, 100]},
    {"id": 3, "atom": "[Ph]", "point_2d": [50, 100]}
  ],
  "bonds": [
    {"atom1": 0, "atom2": 1, "bond_type": "double"},
    {"atom1": 0, "atom2": 2, "bond_type": "single"},
    {"atom1": 2, "atom2": 3, "bond_type": "single"}
  ]
}
\end{lstlisting}
}
\end{AIbox}
\caption{Prompt template for the simplified graph.}
\label{fig:prompt_for_simplified_graph}
\end{figure*}


\begin{figure*}[!ht]
\begin{AIbox}{Prompt template for the graph}
    {
You are an expert model specializing in the analysis of chemical diagrams. Your sole task is to view the third (last) provided image and convert the molecular structure into a graph representation, consisting of atoms and bonds.

Before you begin analyzing the target molecular structure, please carefully examine the first image provided (Figure~\ref{fig:bond_examplar}), which serves as a visual reference example. 
This example demonstrates the different types of chemical bonds and their visual representations. 
Use this visual guide to accurately identify and classify bond types in the structure you will analyze. 

Present the results as a single, complete JSON object.

Follow these instructions meticulously:

\begin{itemize}[leftmargin=*, label=\textbullet, itemsep=2pt]
\item JSON Structure: The final JSON object must contain exactly two top-level keys:
\begin{itemize}
    \item   \texttt{"atoms"}: A list of objects, where each object represents one atom.
    \item   \texttt{"bonds"}: A list of objects, where each object represents one bond.
    \item   \texttt{"brackets"}: A list of objects, where each object represents one bracket.
\end{itemize}
\item   Atom Attributes: Each object in the  \texttt{"atoms"} list must contain the following keys:
\begin{itemize}
    \item    \texttt{"id"}: A unique integer identifier for the atom, starting from 0.
    \item    \texttt{"atom"}: The atom's symbol (e.g., C, O, N). If an abbreviation or group is shown (e.g., Ph, Boc), treat it as a single unit and represent it as a string (e.g., [Ph]).
    \item   \texttt{"point\_2d"} : The [x, y] coordinates of the atom in the image.
    \item   \texttt{"charge"}: (Optional) The formal charge of the atom. An integer (e.g., 1, -1). Only include this key if a charge is explicitly shown.
    \item \texttt{"isotope"}: (Optional) The mass number of an isotope. An integer (e.g., 14 for \ce{^{14}C}). Only include this key if an isotope is explicitly shown.
    \item   \texttt{"valence"}: (Optional) The explicit valence state of the atom. An integer. Only include this key if the valence is explicitly specified (which is rare).
    \item   \texttt{"radical"}: (Optional) The radical state of the atom. Use 1 for a doublet (single radical electron), 2 for a singlet (carbene/nitrene), or 3 for a triplet. Only include this key if a radical is explicitly shown.
\end{itemize}
\item   Bond Attributes: Each object in the \texttt{"bonds"} list must contain three keys:
\begin{itemize}
    \item   \texttt{"atom1"} and \texttt{"atom2"}: The integer ids of the two atoms connected by the bond.
    \item   \texttt{"bond\_type"}: The type of the bond. The value must be one of the following strings. Refer to the visual example image in (Figure~\ref{fig:bond_examplar}) to understand how each bond type appears:
    \begin{itemize}
        \item   \texttt{"single"}: A single bond (single line)
        \item   \texttt{"double"}: A double bond (double line)
        \item   \texttt{"triple"}: A triple bond (triple line)
        \item   \texttt{"aromatic"}: Aromatic bond
        \item   \texttt{"solid wedge"}: Solid wedge bond (thick line indicating atom pointing out of plane)
        \item   \texttt{"dashed wedge"}: Dashed wedge bond (dashed line indicating atom pointing into plane)
        \item   \texttt{"hollow wedge"}: Hollow wedge bond (open triangle indicating atom pointing out of plane)
        \item   \texttt{"wavy"}: Wavy bond (indicating unknown or unspecified stereochemistry)
        \item   \texttt{"any"}: Any bond type (often depicted as a simple, non-stereochemical dashed line, typically for query structures)
        \item   \texttt{"bold"}: Bold bond (thick line, often indicating bonds in the plane of the page)
        \item   \texttt{"dashed bold"}: Dashed bold bond
        \item   \texttt{"dashed double"}: Dashed double bond
        \item   \texttt{"dashed triple"}: Dashed triple bond
        \item   \texttt{"single or double"}: single or double bond represented by triple dashed line
        \item   \texttt{"bold double"}: Bold double bond
    \end{itemize}
\end{itemize}
\end{itemize}

}
\end{AIbox}
\end{figure*}

\begin{figure*}[!ht]
\begin{AIbox}{Prompt template for the graph}
\begin{itemize}[label={}]
    \item
    \begin{itemize}[label={}]
        \item
        \begin{itemize}
            \item   \texttt{"double either"}: Double bond with unknown E/Z configuration
            \item   \texttt{"single or aromatic"}: Single or aromatic bond (ambiguous)
            \item   \texttt{"double or aromatic"}: Double or aromatic bond (ambiguous)
            \item   \texttt{"dative"} - dative bond
            \item   \texttt{"dashed dative"} - Dashed dative bond
            \item   \texttt{"hydrogen"} - Hydrogen bond
            \item   \texttt{"attachment point"} - Attachment point (often used in query structures)
            \item   \texttt{"triple with single dash"} - Triple bond with a single dash
        \end{itemize}
    \end{itemize}
\end{itemize}
\begin{itemize}[leftmargin=*, label=\textbullet, itemsep=2pt]
 \item Bracket Attributes: Each object in the \texttt{"brackets"} list must contain two keys: 
  \begin{itemize}
      \item \texttt{"atoms"}: A list containing the integer ids of all the atoms within a bracket.
     \item \texttt{"mark"}: A number, character or string written in the lower right corner of a bracket indicates the number of times the atoms within the brackets are repeated.
  \end{itemize}
   \item Strict Formatting: You must output only the single JSON object and nothing else. Do not include any additional text, explanations, or formatting like markdown code fences.
\end{itemize}

Example of the required output format:
\begin{lstlisting}[style=json]
{
  "atoms": [
    {"id": 0, "atom": "C", "point_2d": [151, 202]},
    {"id": 1, "atom": "O", "point_2d": [255, 221], "charge": -1},
    {"id": 2, "atom": "N", "point_2d": [132, 434]},
    {"id": 3, "atom": "[Ph]", "point_2d": [59, 100]},
    {"id": 4, "atom": "C", "point_2d": [276, 348], "isotope": 14}
  ],
  "bonds": [
    {"atom1": 0, "atom2": 1, "bond_type": "double"},
    {"atom1": 0, "atom2": 2, "bond_type": "single"},
    {"atom1": 2, "atom2": 3, "bond_type": "wavy"},
    {"atom1": 0, "atom2": 4, "bond_type": "solid_wedge"}
  ],
  "brackets": [
    {"atoms":[0,1,2], "mark": "3"},
    {"atoms":[6,9,10], "mark": "n"},
    {"atoms":[3,5,13], "mark": "n=1,2"},
  ]
}
\end{lstlisting}

Important Notes:
\begin{itemize}
\item The first image you see is the visual example that illustrates different bond types and their visual representations. Study it carefully before analyzing the target structure.
\item The third (last) image contains the chemical structure you need to analyze and convert to the graph representation.
\item If you encounter non-standard styles (one or more) in the test image, please convert them to standardized chemical molecular formula styles for parsing according to the transformation format shown in the second image (Figure~\ref{fig:case_examplar}). This includes 12 common transformation cases.
\end{itemize}

\begin{itemize}
    
\item Case 1: When an asterisk (*) or a curved arc extends from an atom in the molecular structure, it indicates an undefined attachment point. Such cases should be standardized as an R-group variable, denoted as R$\alpha$, R$\beta$, etc., connected to the corresponding atom by a single bond.

\end{itemize}

\end{AIbox}

\end{figure*}

\begin{figure*}[t!]
\begin{AIbox}{Prompt template for the graph}

\begin{itemize}

\item Case 2: If the molecular structure contains a bracket with a subscript (e.g., \texttt{m}) indicating that the atoms inside the bracket are repeated \texttt{m} times, you should first predict the structure as if there were no bracket. In addition, extract the information from the bracket. For example, if the atoms inside the bracket are \texttt{O}, \texttt{C}, and \texttt{N}, and their indices are \texttt{1}, \texttt{2}, and \texttt{3}, then output: \texttt{"brackets": [{"atoms":[1,2,3], "mark": "m"}]}.

\item Case 3: For a metal complex, if the charge is placed outside the square brackets \texttt{[]}, this formal charge should be assigned to the central metal atom. For example, in Case 3, the \texttt{+} charge outside the brackets should be marked as a \texttt{+1} charge on the central \texttt{Ir} atom.

\item Case 4: When an \texttt{HO-} group is connected to a ring via a single bond with an undefined attachment point, the the other end of the bond should be represented as a wildcard \texttt{?}.

\item Case 5: If a bond is truncated at the edge of an image, its other end should be represented as a wildcard atom \texttt{?}.

\item Case 6: A dashed line (representing any bond type) extending from the main structure indicates a bond to an undefined group. The other end of this bond should be treated as an R-group variable, denoted as R$\alpha$.

\item Case 7: When a molecule is connected to a complex graphical structure, that structure should be represented as GROUP$\alpha$. If multiple distinct groups exist, they should be sequentially labeled GROUP$\beta$, GROUP$\gamma$, etc.

\item Case 8: When a negative charge is drawn in the center of a ring (as in a cyclopentadienyl anion), it can be assigned to any atom on the ring. In addition, incomplete arcs within a ring should be interpreted as aromatic bonds.

\item Case 9: When a Markush structure (e.g., \texttt{R}) is connected to a ring via a bond with an undefined attachment point, an \texttt{R} group should be added to every carbon atom on the ring that still has an attached hydrogen. These groups should be labeled sequentially as R$\alpha$, R$\beta$, R$\gamma$, etc., using unique Greek letter subscripts.

\item Case 10: In generic structures or reaction schemes, simple geometric shapes (e.g., circles, ovals, stars, squares) can represent Markush structures. They should be represented as R$\alpha$, R$\beta$, R$\gamma$, etc. All instances of the same geometric shape should share the same label, while different shapes should be assigned unique labels.
    
\item Case 11: The model must ignore non-chemical information, such as background colors, highlights, or decorative overlays on atoms and bonds. It should only recognize the underlying chemical entity (e.g., the \texttt{F} atom, not the red oval on top of it).

\item Case 12: A label like \texttt{d\_n} or \texttt{dn} indicates that \texttt{n} hydrogen atoms on an aromatic ring have been replaced by deuterium (D) atoms. These deuterium (D) atoms should be explicitly represented in the structure.

\end{itemize}

\end{AIbox}
\caption{
  Illustration of the prompt template for the graph generation task.
  This template adopts a few-shot, in-context learning structure. It is composed of two visual exemplars, namely a bond example (Figure~\ref{fig:bond_examplar}) and a case example (Figure~\ref{fig:case_examplar}), followed by the query, which is the chemical formula of the target molecule. The entire sequence is then fed into the model as a single prompt.
}
\label{fig:prompt_for_graph}
\end{figure*}

\section{Annotation Details}
\label{apx:e}
We recruited a total of 47 professional chemical annotators to participate in our chemical reaction annotation task. All annotators possess backgrounds in chemistry-related disciplines, including 9 master's degree candidates and 31 bachelor's degree holders. Their academic specialties cover multiple chemistry-related fields such as Chemistry, Applied Chemistry, Materials Chemistry, Chemical Engineering and Technology, and Organic Chemistry. Prior to formally commencing the annotation work, all annotators passed rigorous chemical knowledge assessments and comprehensive annotation protocol training. This ensures their solid foundation in chemistry and consistent adherence to annotation standards, thereby guaranteeing the professionalism and reliability of the annotated data.
\section{Source Journals of MolRecBench-Wild}
\label{apx:f}
The journals selected for data collection and the number of molecular images obtained from each journal are shown in Tab.\ref{tab:journal_sources}.

\begin{table*}[h]
\centering
\caption{Journal Sources and the Number of Molecules Extracted from each.}
\small
\setlength{\tabcolsep}{5pt} 
\begin{tabular}{c c c}  
\toprule
\textbf{ID} & \textbf{Journal Name}                              & \textbf{ \# Molecules} \\ \midrule
1            & \raggedright Angewandte Chemie International Edition  & 1505                                      \\ 
2            & \raggedright Accounts of Chemical Research            & 164                                       \\ 
3            & \raggedright ACS Catalysis                           & 587                                       \\ 
4            & \raggedright ACS Central Science                     & 464                                       \\ 
5            & \raggedright Journal of the American Chemical Society & 1046                                      \\ 
6            & \raggedright Nature Chemistry                        & 100                                       \\ 
7            & \raggedright Chemical Science                        & 1198                                      \\ 
\bottomrule
\end{tabular}
\label{tab:journal_sources}
\end{table*}

\end{document}